\documentclass{article}
\usepackage{ijcai24}
 
\usepackage[hyphens]{url}  
\usepackage{graphicx} 
\usepackage{amsmath}  
\usepackage{times}
\usepackage{soul}
\usepackage{url}
\usepackage[colorlinks,citecolor=blue]{hyperref}
\usepackage[utf8]{inputenc}
\usepackage[small]{caption}
\usepackage{graphicx}
\usepackage{amsmath}
\usepackage{amsthm}
\usepackage{booktabs}
\usepackage{algorithm}
\usepackage{algorithmic}
\usepackage[switch]{lineno}
\usepackage{multirow} 
\usepackage{graphicx}
\usepackage{makecell}
\usepackage{graphicx}
\usepackage{mathtools}
\usepackage{amssymb} 
\usepackage{tablefootnote}  
\usepackage[hang,flushmargin]{footmisc}  
\usepackage[flushleft]{threeparttable}  
\usepackage{tabularx}
\usepackage[table,xcdraw]{xcolor}
\usepackage{longtable}

\usepackage{acro}
\DeclareRobustCommand{\acrodef}[2]{\DeclareAcronym{#1}{short=#1,long=#2}}

\acrodef{ICL}{In-context learning}

\title{\includegraphics[scale=0.02]{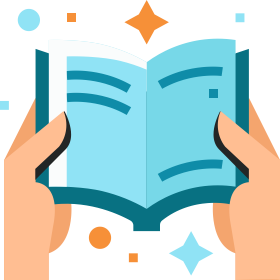} Grimoire is All You Need for Enhancing Large Language Models}

\author{
Ding Chen$^{1,}$\footnote{Equal contribution.}
\and
Shichao Song$^{2,*}$\and
Qingchen Yu$^{3}$\and
Zhiyu Li$^{3,}$\footnote{Corresponding author:lizy@iaar.ac.cn} \and
Wenjin Wang$^{3}$ \and
Feiyu Xiong$^{3}$ \And
Bo Tang$^{3}$
\affiliations
$^1$Beihang University\\
$^2$Renmin University of China\\
$^3$Institute for Advanced Algorithms Research, Shanghai\\
\emails
}

\begin{document}
\maketitle

\begin{abstract}

\ac{ICL} is one of the key methods for enhancing the performance of large language models on specific tasks by providing a set of few-shot examples. However, the \ac{ICL} capability of different types of models shows significant variation due to factors such as model architecture, volume of learning data, and the size of parameters. Generally, the larger the model's parameter size and the more extensive the learning data, the stronger its \ac{ICL} capability. In this paper, we propose a method \textsc{SleIcl} that involves learning from examples using strong language models and then summarizing and transferring these learned skills to weak language models for inference and application. This ensures the stability and effectiveness of \ac{ICL}. Compared to directly enabling weak language models to learn from prompt examples, \textsc{SleIcl} reduces the difficulty of \ac{ICL} for these models. Our experiments, conducted on up to eight datasets with five language models, demonstrate that weak language models achieve consistent improvement over their own zero-shot or few-shot capabilities using the \textsc{SleIcl} method. Some weak language models even surpass the performance of GPT4-1106-preview (zero-shot) with the aid of \textsc{SleIcl}\footnote{The source code is available at GitHub:\url{https://github.com/IAAR-Shanghai/Grimoire}}. 

\end{abstract}

\section{Introduction}

\begin{figure}[!t]
    \centering
    \includegraphics[width=1.0\linewidth]{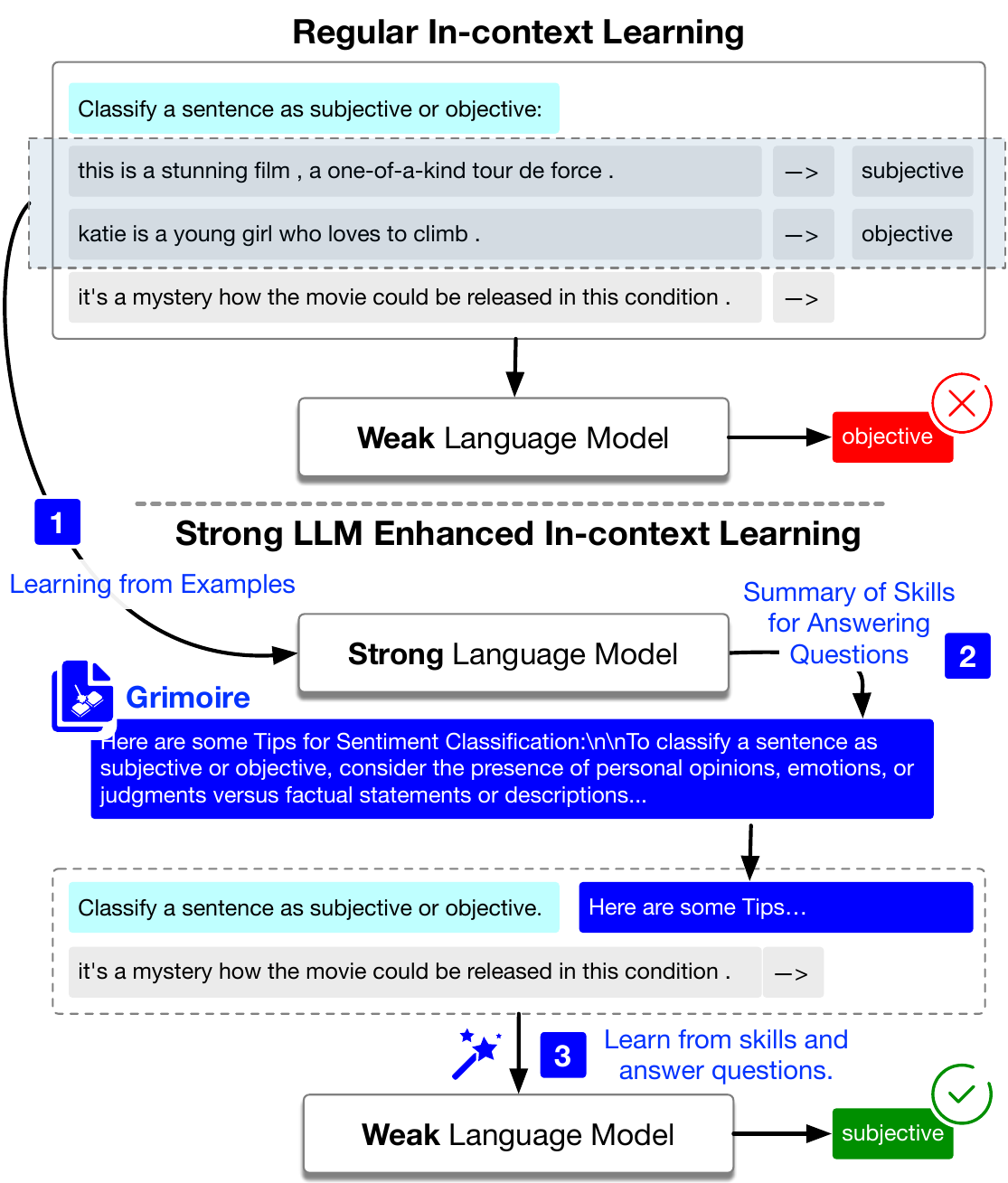}
    \caption{Compared to having a language model directly engage in Regular In-Context Learning (Regular ICL), Strong LLM Enhanced In-Context Learning (\textsc{SleIcl}) involves having a strong language model initially learn and summarize techniques based on representative samples. Subsequently, the generated techniques (grimoire) are incorporated as part of the prompt to guide the weak language models in their responses. }
    \label{fig:intro_example}
\end{figure}

The continuous evolution of large language models has positioned \acf{ICL} prominently in foundational capabilities, attracting significant research interest~\cite{SurveyICL_23_arXiv_PKU}. \ac{ICL} improves model performance in novel tasks and enhances task-specific outcomes by providing a concise set of prompt examples within a given task context. In contrast to parameter-based optimization methods like supervised fine-tuning, \ac{ICL} eliminates the need for extensive pre-use training or updates to the original large Language models, thus offering broader applicability. In a range of natural language processing tasks—such as sentiment classification~\cite{SST2&SST5_13_CEMNLP_Stanford}, topic categorization~\cite{AgNews&DBPedia_15_NIPS_NYU}, and natural language inference~\cite{RTE_05_MLCW_BIU}, the provision of well-crafted demonstration examples to less powerful models (e.g., those with 7B or 13B parameters) often results in performance that matches or surpasses more advanced models (such as GPT-4)~\cite{GlobalE_22_ACL_UCL,IDS_23_arXiv_NTU}. This phenomenon significantly fuels researchers' interest in exploring the potential of \ac{ICL}.

Prior research on \ac{ICL} has emphasized the importance of selecting and utilizing demonstration examples~\cite{KATE_22_DeeLIO_Duke}. Studies suggest that the number, quality, and relevance of these demonstration examples, along with their sequential ordering in the prompt queue, markedly affect the efficacy of ICL in LLMs~\cite{Few-shot_22_NeurIPS_OpenAI,ICL-TC_23_ACL_McGill}. This phenomenon stems from the diverse learning and knowledge transfer capabilities inherent in models of varied scales and architectures. For instance, given identical input example pairs $(x_1, y_1), (x_2, y_2), (x_3, y_3)$, weak models frequently demonstrate a reduced proficiency in approximating $y$ compared to their more advanced counterparts. Consequently, rather than relying on weak models to directly assimilate these examples, a more efficacious strategy may involve harnessing the learning prowess of stronger models. This can be achieved by isolating the process of fitting demonstration examples and utilizing the empirical function $f(x) \rightarrow y$, inferred from more stronger models, to direct the learning in weaker models. Ultimately, this approach serves to diminish the complexity involved in direct learning from demonstrations, thus bolstering the performance of weak models.
 
As depicted in Figure \ref{fig:intro_example}, this paper presents a new paradigm for augmenting \ac{ICL}, denominated as \textsc{SleIcl} (\textbf{S}trong \textbf{L}LM \textbf{E}nhanced ICL). This method exploits the superior capabilities of strong LLM to assimilate functions or problem-solving skills from demonstration examples, metaphorically termed as a \textbf{Grimoire}. Once the grimoire is generated by the strong LLM, it serves as a substitute for the original prompt examples, thus streamlining the learning process for weak models. In conventional \ac{ICL} methodologies, LLMs generally must navigate through a collection of demonstration examples for each query~\cite{EPR_22_arXiv_TelAvivU}, choosing the most suitable examples to optimize performance. Conversely, \textsc{SleIcl} necessitates only a solitary instance of example selection for a designated task. Once the grimoire is produced, it can direct weak models in addressing queries within that task, yielding results that exceed those achievable through regular \ac{ICL}.

\section{Related Works}

\subsection{In-context Learning of Large Language Model}

\ac{ICL} has emerged as a novel learning paradigm, initially proposed and applied in the pre-training of GPT-3~\cite{Few-shot_22_NeurIPS_OpenAI}. \ac{ICL} can efficiently learn tasks with a small number of prompt examples, without the parameter updates. Why \ac{ICL} is effective has sparked widespread discussion. ~\cite{DDPD_22_NeurIPS_DeepMind} suggests that the \ac{ICL} capability of models is driven by the distributional characteristics of the data itself, emphasizing the importance of data structure. ~\cite{GINC_22_arXiv_Stanford} suggests that contextual learning takes place as the language model infers shared latent concepts among examples within prompts. \cite{ICL-SoSFC_22_NeurIPS_Stanford} indicates that models can learn specific functions based on encoded prompt samples, achieving performance comparable to specific task learning algorithms. Combining the above works, we find that the \ac{ICL} capabilities of models are more derived from learning the distributional features of example samples or underlying rules, rather than necessarily relying on the specific examples. Moreover, the greater the parameter size of large language models, the more robust their corresponding \ac{ICL} capabilities~\cite{Few-shot_22_NeurIPS_OpenAI,GLER_22_EMNLP_SNU}, establishing the theoretical foundation for our work.

\begin{figure*}[htp]
    \centering
    \includegraphics[width=1.0\linewidth]{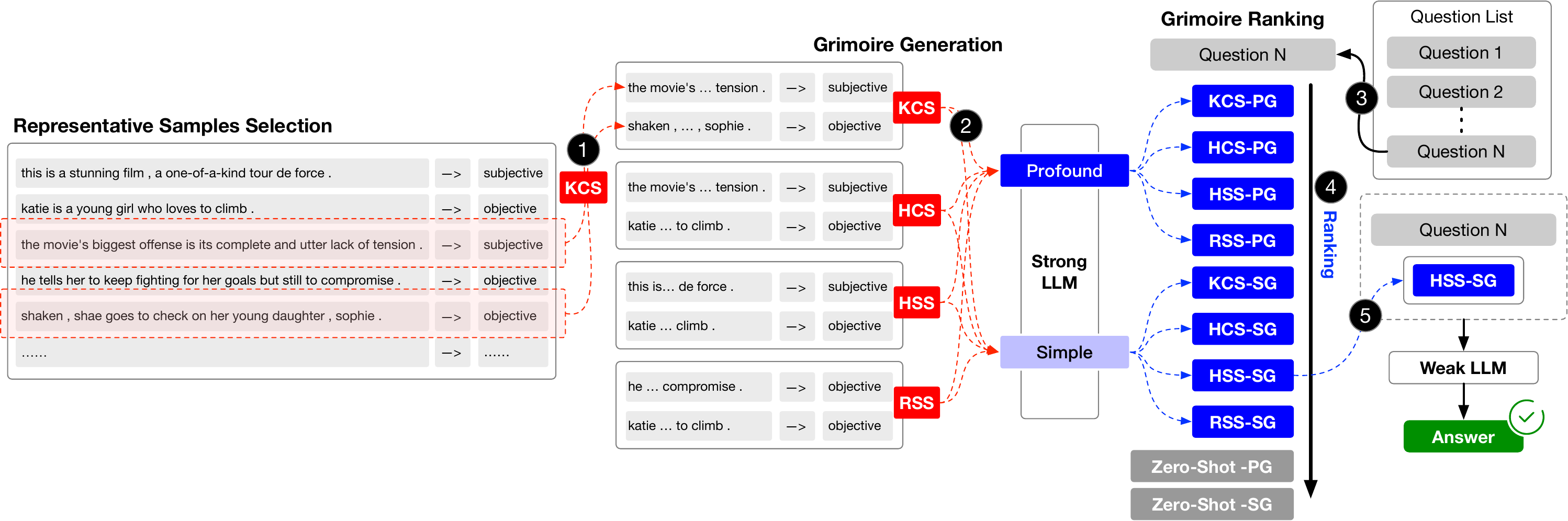}
    \caption{Framework of proposed \textsc{SleIcl} method. First, multiple sets of representative samples are obtained using different sample selection methods (KCS, HCS, HSS, RSS), with each set sampled in a stratified manner based on labels. Subsequently, corresponding profound grimoires (PG) and simple grimoires (SG) are generated based on each sample set. Additionally, zero-shot-PG and zero-shot-SG, generated without samples, are included. Finally, all grimoires are ranked based on given test samples, and the optimal grimoire is handed over to the weak LLM for response.} 
    \label{fig:framework}
\end{figure*}

\subsection{Prompt Engineering of Demo Examples}
Extensive research indicates that the construction of demonstration examples is crucial for the performance of \ac{ICL}~\cite{GLER_22_EMNLP_SNU,IDS_23_arXiv_NTU}. Furthermore, recent research has enhanced \ac{ICL} performance by optimizing both the characteristics of demonstration examples and the order and selection strategies of example samples.
 
\textbf{Characteristics of Demonstration Examples}. The study by ~\cite{Rethinking_22_EMNLP_UW} highlights that the influence of prompt samples on \ac{ICL} performance is attributable to four principal elements: input-label mappings, label space, and sample distribution. Nonetheless, it is notable that opinions diverge concerning the effect of input-label mappings relationships—namely, label accuracy—on \ac{ICL} performance. Certain studies propose that input-label mappings relationships potentially influence \ac{ICL} performance, with ~\cite{ICUL_23_arXiv_Harvard} observing that label inversion within contextual examples markedly impacts the outputs of the Bloom model. In contrast, findings by ~\cite{SUL-ICL_23_arXiv_Google} demonstrate that larger models generally exhibit more robust \ac{ICL} capabilities and excel in deciphering label mapping relationships from contextual examples relative to their smaller counterparts.

\textbf{Ordering of Demonstration Examples}. The ordering of samples represents a critical aspect of prompt sample construction, with various sorting methods leading to differences in \ac{ICL} performance~\cite{Calibrate_21_PMLR_UCB}. The study by~\cite{KATE_22_DeeLIO_Duke} introduced KATE, a system that selects demonstration examples based on semantic similarity. Expanding on this, they examined various ordering methods and found that their impact on KATE's performance was minimal. Research from \cite{GlobalE_22_ACL_UCL} employed GlobalE and LocalE to sequence demonstration examples and uncovered a positive relationship between information entropy and \ac{ICL} performance. As the parameter scale increases, models become more proficient at \ac{ICL}, and their sensitivity to sample ordering diminishes accordingly~\cite{ICL-TC_23_ACL_McGill}.

\textbf{Selecting Demonstration Examples}. Selection of demonstration examples is a pivotal stage in crafting prompt samples, with a substantial effect on \ac{ICL} performance~\cite{KATE_22_DeeLIO_Duke}. Presently, selection of demonstration examples methodologies are primarily categorized into three types: semantic similarity~\cite{KATE_22_DeeLIO_Duke,Vote-k_22_arXiv_HKU}, clustering~\cite{AutoCoT_23_ICLR_SJTU}, and information entropy~\cite{CBSMaxIG_23_arXiv_NUS}. Moreover, certain studies have introduced specialized models to score demonstration examples, aiming to select representative demonstration examples~\cite{EPR_22_arXiv_TelAvivU,Self-AdaptiveICL_23_ACL_SHLab,LENS_23_EMNLP_FDU}. Previous research on the selection of demonstration examples can be categorized based on the granularity of selection. One strategy involves obtaining instance-level examples, where retrieval is performed for every test query~\cite{KATE_22_DeeLIO_Duke,Vote-k_22_arXiv_HKU,EPR_22_arXiv_TelAvivU,Self-AdaptiveICL_23_ACL_SHLab}. Alternatively, task-level example retrieval is utilized as prompts for all test samples~\cite{AES-RL_22_EMNLP_Uchicago,CBSMaxIG_23_arXiv_NUS,LENS_23_EMNLP_FDU}, which is less resource-intensive compared to instance-level retrieval. However, such selected samples may not be sufficiently representative or could be substandard, potentially resulting in modest improvements in \ac{ICL} performance or reduced stability. Researchers and practitioners must deliberate these aspects when advancing and implementing \ac{ICL} methodologies to ensure the most effective deployment of demonstration examples.

\section{Enhancing LLMs with Grimoire}

\subsection{Problem Formulation}

In the ICL scenario, the key is to find suitable demonstration examples to achieve higher prediction accuracy. Conversely, in the \textsc{SleIcl} scenario, the focus is on finding an appropriate grimoire. Specifically, given a task $\mathcal{T}$ that needs to be solved by a weak language model $\mathcal{W}$, along with a training set $T_{\mathcal{T}}$ and a validation set $T_{\mathcal{D}}$, our objective is to identify the optimal grimoire $g_i \in \mathcal{G}$ produced by the strong language model $\mathcal{L}$ that enhances the performance of $\mathcal{W}$ on $T_{\mathcal{D}}$ compared to ICL prompting $\mathcal{W}$ with n-shot examples, $\text{S}_n \in {T_{\mathcal{T}}}$:

\begin{equation}
    \begin{cases}
        g^*=\arg \max_{g_{i} \in \mathcal{G} } \textsc{SleIcl} \left ( \mathcal{W}_{T_{\mathcal{D}}} | g_i \right ) \\
        \text{S}^*_n=\arg \max_{\text{S}_n \in {T_{\mathcal{T}}}} \text{ICL} \left (\mathcal{W}_{T_{\mathcal{D}}}|\text{S}_n\right ) \\
        \textsc{SleIcl} \left (\mathcal{W}_{T_{\mathcal{D}}}|g^*\right )  > \text{ICL} \left (\mathcal{W}_{T_{\mathcal{D}}}|\text{S}^*_n\right )
    \end{cases}
    \label{eq:formulation}
\end{equation}

In Equation~\ref{eq:formulation}, $\textsc{SleIcl} (\cdot)$ denotes prompting with grimoire, and $\text{ICL} (\cdot)$ denotes \ac{ICL} prompting. Figure \ref{fig:framework} shows the enhancing framework based on strong language model $\mathcal{L}$. Initially, we select a representative set of examples $\text{S}_n$ from the training set  $T_{\mathcal{T}}$ provided by the task. In this step, we designed four different selection methods $\text{S}_n \in \left \{ \text{KCS}(n),\text{HCS}(n),\text{HSS}(n),\text{RSS}(n) \right \}$ to find better demonstration examples. Subsequently, even among weak language models, there are variations in learning capabilities. Hence, we have specifically designed two distinct paradigms for generating grimoires, namely the Profound Grimoire (PG) and the Simple Grimoire (SG). Ultimately, by combining four example selection methods with the two grimoire generation strategies, we are able to create a candidate set comprising eight different grimoires for a specific task. In addition, we also added PG and SG generated without samples(zero-shot). In the final stage of task evaluation, we implement a grimoire ranking algorithm. This algorithm is designed to select the potentially optimal grimoire for enhancement, corresponding to different problems, thereby further improving the performance of the weak language models.

\subsection{Representative Samples Selection}

This process involves choosing a subset of data that captures the underlying patterns and complexities of the entire samples, aiming to improve the efficiency and effectiveness of grimoire generation. Various methods have been developed to tackle this challenge, each with its own methodology and focus.

\subsubsection{K-means Clustering Selection (KCS)}
K-means Clustering Selection refers to the process of using the k-means algorithm to cluster the semantic representations of a sample set and selecting the nearest $n$ samples to each of the $k$ cluster centers as representative samples. Consequently, a collection of $n*k$ exemplary sample sets is obtained. We believe that a diverse set of representative samples may potentially enhance the performance of large models~\cite{AutoCoT_23_ICLR_SJTU}, and is more conducive to allowing strong models to generalize answer skills on a holistic level. This, in turn, increases the universality of the final generated grimoire without resulting in localized optimization. In KCS, the hyperparameters include the number of clusters $k$, and the number of samples $n$ in each cluster.
 
\subsubsection{Hierarchical Clustering Selection (HCS)}
The Hierarchical Clustering Selection method employs the hierarchical clustering algorithm to perform a detailed hierarchical clustering analysis of the sample set. HCS selects representative samples from the cluster centers identified at various levels in a dendrogram, aiming to capture and display the rich hierarchical semantic features within the samples. HCS not only reveals subtle associations between samples, but also has the advantage of not requiring a predefined number of clusters, making HCS more advantageous when dealing with datasets that have complex semantic structures. Moreover, by providing multi-level semantic feature representations, HCS adds a more diversified basis for choices to the subsequent grimoire ranking algorithms, thus allowing a more precise reflection of the true semantic relationships between samples, and further optimizing the effectiveness and quality of ranking.

\subsubsection{Hard Samples Selection (HSS)}
Hard Samples Selection refers to selecting samples that are easily mispredicted by weak models as representative samples. Hard examples may contain information and knowledge that are either lacking or only partially understood by weak models in solving a given task. Thus, we aim to effectively compensate for the deficiencies and insufficiencies of weak language models in addressing specific problems by using strong language models to extract and refine the skills needed to solve these hard examples. Our first step involves conducting zero-shot testing on the training sets using the weak model, and recording the predicted labels for each sample. When we perform hard sample selection, we choose a fixed proportion of hard samples based on a given ratio. For instance, when the ratio is set to 0.3, it means selecting 30\% of samples from the mispredicted examples and the remaining 70\% from the correctly predicted ones.

\subsubsection{Random Samples Selection (RSS)}
Random Samples Selection is a method that selects representative samples from a dataset in an non-discriminatory manner. Samples are picked entirely at random. This approach is beneficial for maintaining an impartial sample distribution, particularly when little is known about the datasets structure or when seeking a baseline method for comparison with more complex selection methods like KCS, HCS, and HSS. RSS’s simplicity makes it efficient for large datasets and useful for preliminary explorations or alongside other selection methods.

\subsection{Grimoire Generation}
Upon completing the selection of representative examples, it becomes imperative to employ strong language models to generate content aimed at guiding weak language models in responding. This generated content has been termed "Grimoire". However, given the substantial variations in \ac{ICL} capabilities among language models of different parameter sizes, employing a singular grimoire generation approach for all weak language models presents a challenge. Consequently, we have devised two fundamental types of grimoire generation paradigms: Profound Grimoire Generation and Simple Grimoire Generation. Our hypothesis posits that for larger models with enhanced reasoning and comprehension skills, Profound grimoire tends to yield superior outcomes through the use of detailed skill explanations and the generation of diverse answers based on reference samples. In contrast, weake language models are more likely to attain improved results with more concise grimoires featuring straightforward examples.

\begin{figure}[!t]
    \centering
    \includegraphics[width=0.9\linewidth]{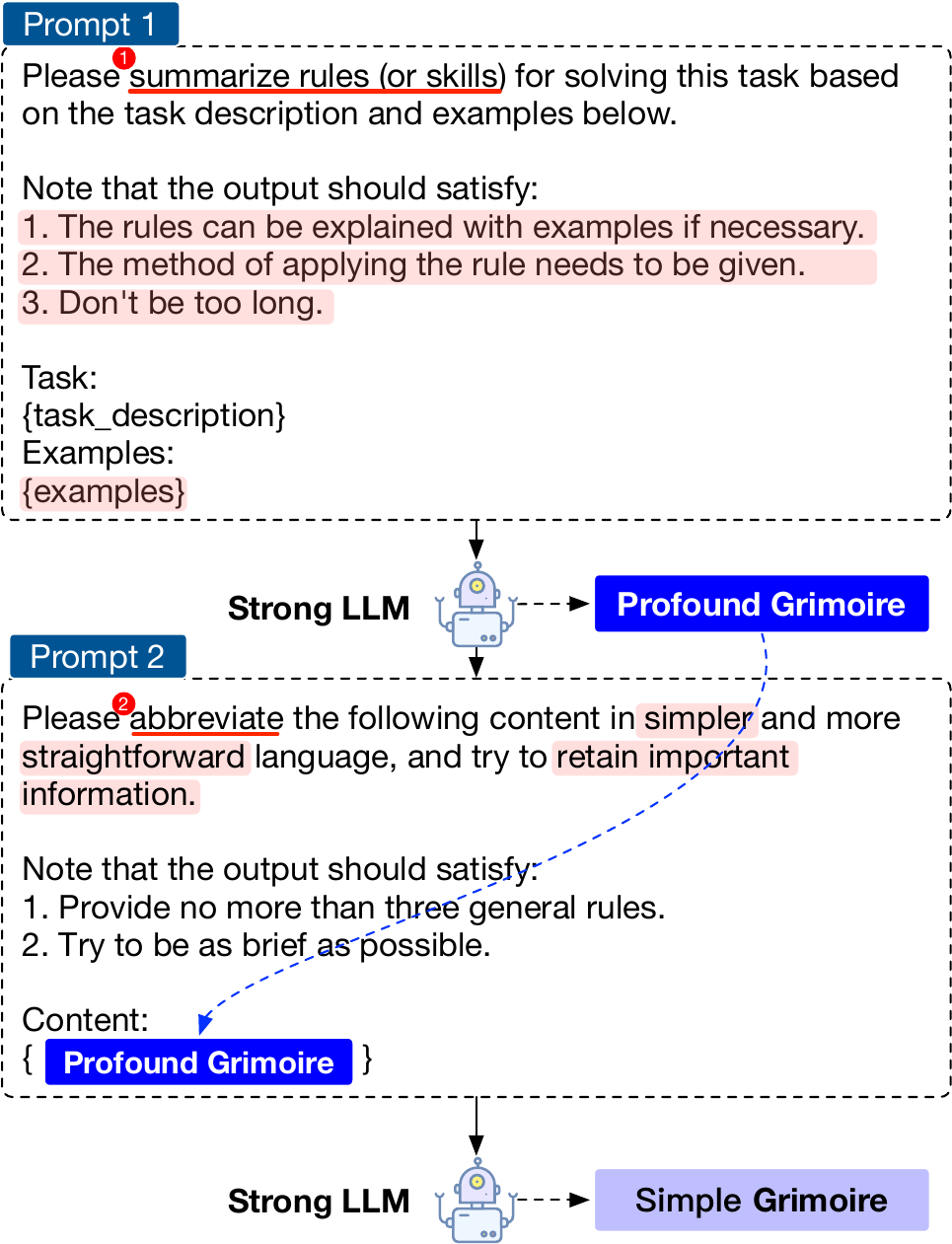}
    \caption{Workflow for grimoire generation.}
    \label{fig:prompt_template}
\end{figure}

\subsubsection{Profound Grimoire (PG)}

The profound grimoire is characterized by an abundance of information and details, necessitating that the guided weak language model possesses robust \ac{ICL} abilities. Consequently, a specialized prompt template for generating PG has been developed, as illustrated in Figure \ref{fig:prompt_template}. The essence of this prompt template involves compelling the strong language model to synthesize the skills essential for solving the given task, drawing upon the provided task description and representative samples. We anticipate utilizing the strong language model's exceptional summarization and abstraction capabilities to distill universal solutions for specific tasks, providing explanations alongside the given samples, as well as methods for skill application, mirroring the instructional approach of human teachers. This facilitates the weak language model's comprehension of task requirements and accelerates its learning of the necessary skills for task resolution. The ultimate stipulation in the template is to constrain the length of the strong language model's output, thereby preventing the grimoire from surpassing the weak language model's context capacity, as an excessively extended grimoire could yield detrimental outcomes.

\subsubsection{Simple Grimoire (SG)}
 
The simple grimoire represents a simplified version of the PG. Crafted for greater conciseness and clarity, it retains critical information from the PG, thereby enabling language models with lesser \ac{ICL} capability to comprehend the grimoire and distill the essential information and methodologies for addressing specific tasks (Template illustrated in Figure \ref{fig:prompt_template}). The prompt template for generating SG was not reconfigured to maintain the uniformity in content and skill articulation between SG and PG. Employing a simplified methodology guarantees robust consistency between these two grimoires for the identical task. The benefit of this consistency lies in the fact that when comparing the impacts of both grimoires on the same weak language model, it becomes clear that any efficacy disparities are primarily attributed to the complexity of the articulation, rather than variances in content and meaning. This approach aids users in selecting the more appropriate grimoire when the suitability for a specific weak language model’s comprehension capabilities is ambiguous.

\subsection{Grimoire Ranking} \label{sec:rank}

By integrating four representative samples selection methods (KCS, HCS, HSS and RSS) and zero-shot method with two grimoire generation paradigms (PG and SG), for each task, we are able to obtain ten types of grimoires: 

\begin{equation}
    g_i \in \mathcal{G}  = \left \{ \text{KCS,HCS,HSS,RSS,Zero-Shot}\right \} \times \left \{  \text{PG,SG} \right \} 
    \label{eq:grimoires}
\end{equation}

Among these, the best grimoire will be chosen to serve as the prompt for our \textsc{SleIcl} method. Hence, the formulated objective is to identify the optimal grimoire $g^*$ from the candidate set $\mathcal{G}$ for a given task $q_j$ in the test dataset $T_{\mathcal{D}}$:

\begin{equation}
    \forall q_j \in T_{\mathcal{D}}, \; g^* =\arg \max_{g_{i} \in \mathcal{G} } u \left ( {q_j}, g_i | \mathcal{W} \right )
    \label{eq:objective}
\end{equation}

In Equation~\ref{eq:objective}, $u \left ( {q_j}, g_i | \mathcal{W} \right )$ denotes the utility of grimoire $g_i$ for query $q_j$ on a specific weak language model $\mathcal{W}$. Therefore, the crux of the problem shifts to how to define this utility function. Here, we propose two approaches, similarity-based method and classifier-based method.

\textbf{Similarity-based method}. The simplest way to evaluate utility is to calculate the similarity between query and grimoire. Therefore, we employ an embedding model from the MPNet\cite{mpnet_20_NIPS_Microsoft} to embed both the query and the grimoire. Subsequently, the utility is determined by calculating the cosine similarity as shown in Equation~\ref{eq:sim}. Certainly, this approach solely relies on similarity, disregarding critical features like task type and grimoire type. Consequently, it cannot effectively model the utility function. Hence, we opt to pre-train a deep neural network.

\begin{equation}
    u_\text{sim}(\cdot) \triangleq \cos(\text{embed}{g_i},\text{embed}{q_j})
    \label{eq:sim}
\end{equation}

\textbf{Dual-tower deep neural network classifier method}. This method is composed of multiple layers of neural networks, as illustrated in simplified form by Equation~\ref{eq:tower}. The $\text{dense}(\cdot)$ portion consists of a two-layer densely connected neural network with residual connections. Then, a self-attention mechanism transforms the concatenated query and grimoire, and the resulting joint representation is fed into a classification head with four layers. For detailed feature engineering, neural network architecture and training process, please refer to the appendix.

\begin{equation}
    \begin{cases}
        \text{joint} = \text{dense}{(g_i)} \otimes \text{dense}{(q_j)} \\
        u_\text{tower}(\cdot) \triangleq \text{classifier-head}(\text{self-attn}(\text{joint}))
    \end{cases}
    \label{eq:tower}
\end{equation}

\section{Experiments}

\subsection{Datasets}

Our evaluation encompassed eight datasets across four task categories: \textbf{Sentiment Analysis} (SST5 \cite{SST2&SST5_13_CEMNLP_Stanford} and Subj \cite{Subj_04_AMACL_Cornell}), \textbf{Topic Classification} (AgNews \cite{AgNews&DBPedia_15_NIPS_NYU} and TREC \cite{TREC1_02_ICCL_UIUC,TREC2_01_ICHLTR_USC}), \textbf{Natural Language Inference} (RTE \cite{RTE_05_MLCW_BIU} and QNLI \cite{QNLI_16_CEMNLP_Stanford}), and \textbf{Hate Speech Detection} (hate\_sp18 \cite{hate-speech18_18_ALW_HSLT} and ethos \cite{ethos_20_arXiv_TAU}). A series of preprocessing activities were conducted on the original data from these datasets to transform all samples into a consistent and unified format, thus enabling smooth subsequent testing. Initially, samples from the original dataset that are excessively lengthy were excluded to prevent the issue of prompts exceeding the context window limits of smaller models during few-shot testing. Subsequently, the original dataset was partitioned into a training set and a test set (unless pre-segmented), with the training set comprising 2000 samples and the test set 1000 samples. In cases where the total number of samples in the original dataset fell below 3000, a distribution ratio exceeding 2:1 was maintained between the training and test sets to ensure sufficient samples for subsequent example instance selection in the prompts.

\begin{table*}[!t]
    \centering
    \resizebox{\linewidth}{!}{
    \begin{threeparttable}
        \caption{Detailed evaluation results of GPT3.5-Turbo enhanced by grimoire.}
        \label{tab:Detailed evaluation results}
        \begin{tabular}{cccccccccccc}
            \toprule
            \multirow{2}{*}[-0.8ex]{Model} & \multirow{2}{*}[-0.8ex]{Submodel} & \multirow{2}{*}[-0.8ex]{n/k-shot} & \multicolumn{2}{c}{Sentiment Analysis} & \multicolumn{2}{c}{Topic Classification} & \multicolumn{2}{c}{\makecell[l]{Natural Language \\ Inference}} & \multicolumn{2}{c}{Hate Speech Detection} & \multirow{2}{*}[-0.8ex]{AVG} \\
           \cmidrule(lr){4-5} \cmidrule(lr){6-7} \cmidrule(lr){8-9} \cmidrule(lr){10-11} 
            & & & SST5 & Subj & AgNews & TREC & RTE & QNLI & hate\_sp18 & ethos & \\
            \midrule
            \multirow{2}{*}{GPT3.5-Turbo} & Zero-Shot & - & 53.87\% & 39.00\% & 83.82\% & 81.69\% & 77.67\% & 67.40\% & 78.76\% & 81.20\% & 70.43\% \\
            & Few-Shot & n=4 & 54.07\% & 43.67\% & \textbf{85.75\%} & 77.76\% & 79.07\% & 74.80\% & 68.31\% & 76.87\% & 70.04\% \\
            \specialrule{0em}{1pt}{1pt}
            \midrule
            \multirow{12}{*}{Single Grimoire (GPT3.5-Turbo)} & KCS-PG & k=4 & 54.04\% & \underline{60.60\%} & 81.87\% & 76.81\% & 80.20\% & 75.50\% & 78.24\% & 84.20\% & \underline{73.93\%} \\
            & KCS-SG & k=4 & 50.23\% & 50.80\% & 84.00\% & 74.47\% & 79.93\% & 73.60\% & 75.23\% & 84.00\% & 71.53\% \\
            & HCS-PG & k=4 & 50.90\% & 43.87\% & 79.53\% & 73.20\% & 74.20\% & 76.27\% & 83.67\% & 81.80\% & 70.43\% \\
            & HCS-SG & k=4 & 50.20\% & 45.20\% & 85.13\% & 80.49\% & \underline{80.80\%} & 75.20\% & 78.89\% & 83.00\% & 72.36\% \\
            & HSS-PG & k=4 (r=1.0) & 51.60\% & \textbf{65.00\%} & 84.80\% & \textbf{83.01\%} & 75.80\% & 74.67\% & 82.98\% & 83.33\% & \textbf{75.15\%} \\
            & HSS-SG & k=4 (r=1.0) & \textbf{55.27\%} & 48.80\% & 83.80\% & \underline{82.43\%} & 80.13\% & 73.07\% & 77.87\% & 84.07\% & 73.18\% \\
            & HSS-PG & k=4 (r=0.5) & 50.97\% & 46.93\% & \underline{85.31\%} & 79.67\% & 78.07\% & 68.60\% & \textbf{86.27\%} & 84.07\% & 72.49\% \\
            & HSS-SG & k=4 (r=0.5) & 53.10\% & 47.80\% & 84.27\% & 72.54\% & \textbf{81.47\%} & 73.33\% & 79.01\% & \textbf{85.33\%} & 72.11\% \\
            & RSS-PG & k=4 & 51.57\% & 52.67\% & 82.47\% & 72.03\% & 80.40\% & \textbf{79.40\%} & \underline{85.34\%} & 84.13\% & 73.50\% \\
            & RSS-SG & k=4 & \underline{54.30\%} & 51.13\% & 82.53\% & 75.53\% & 78.47\% & \underline{77.00\%} & 75.50\% & 83.20\% & 72.21\% \\
            & Zero-Shot-PG & - & 49.50\% & 57.13\% & 82.31\% & 66.69\% & 76.20\% & 72.33\% & 81.93\% & 84.13\% & 71.28\% \\
            & Zero-Shot-SG & - & 50.67\% & 53.47\% & 82.38\% & 75.10\% & 77.13\% & 74.87\% & 78.77\% & \underline{84.47\%} & 72.11\% \\
            \specialrule{0em}{1pt}{1pt}
            \midrule
            \multirow{2}{*}{\textsc{SleIcl} (GPT3.5-Turbo)} & Similarity-based & - & 52.97\% & 58.53\% & 82.64\% & 78.34\% & 76.60\% & 73.53\% & 79.21\% & 83.87\% & 73.21\% \\
            & Classifier-based & - & 52.23\% & 59.13\% & 83.06\% & 79.03\% & 79.07\% & 74.32\% & 79.93\% & 83.80\% & 73.82\% \\
            \bottomrule
        \end{tabular}
        \begin{tablenotes}
            \footnotesize
            \item \textit{Note}: - indicates that this hyper-parameter is invalid for the current test; n-shot indicates that n samples will be provided for each prediction; k-shot provides a selection of k samples under each label to generate grimoire; r represents the sampling ratio of hard samples. The best performance in each column will be bolded, and the second-best performance will be underlined.
        \end{tablenotes}
    \end{threeparttable}
    }
\end{table*}

\subsection{Configuration and Metrics}
 
\textbf{Large Language models}. We evaluate 6 language models, including GPT3.5-Turbo (175B\footnote{175B is an estimated value.})\footnote{\url{https://openai.com}}, LLaMA2 Chat (70B, 13B)~\cite{Llama2_2023_arXiv_Meta}, Baichuan2 (7B)~\cite{Baichuan2_23_arXiv_Baichuan}, GPT4-1106-preview\footnote{\url{https://openai.com/blog/new-models-and-developer-products-announced-at-devday}}, Phi-2 (2.7B)\footnote{\url{https://www.microsoft.com/en-us/research/blog/phi-2-the-surprising-power-of-small-language-models/}}. Among them, GPT3.5-Turbo and GPT4-1106-review are two important LLMs developed by OpenAI. The LLaMA2 Chat series model is an open-source chat model generated by Meta, and we plan to evaluate the 70B and 13B chat models in this series. Baichuan2 is an open-source LLM developed by Baichuan Inc. , and we plan to evaluate the 7B chat model in this series. Phi-2 is a small language model (SLM) developed by Microsoft. Among these models, GPT4-1106-review will be used as a strong language model for generating grimoires.

\textbf{Baseline}. We designed the following two types baselines to compare with our method: zero-shot and few-shot (n=4).

\textbf{Single Grimoire}. We evaluated the ten types of grimoires in Equation~\ref{eq:grimoires}, where representative samples are stratified by label ($k$ samples per label).

\textbf{\textsc{SleIcl}}. We evaluated the two types of methods proposed in Section~\ref{sec:rank}: similarity-based method and classifier-based method.

\subsection{Performance Analysis}

In Table \ref{tab:Detailed evaluation results}, we present in detail the test results of GPT3.5-Turbo on three types of methods: baseline, single grimoire, and \textsc{SleIcl}. Overall, the single grimoire and \textsc{SleIcl} have better average accuracy than baseline on all task datasets, with HSS-PG (r=0.5) and KCS-PG having the highest average accuracy, exceeding baseline by 4.72\% and 3.5\%, respectively. This indicates that the grimoire generated by strong language model can effectively improve the performance of weak language model on various tasks. And the best performing single grimoire is HSS-PG, and the maximum accuracy on each task dataset is not obviously concentrated on a single grimoire. This indicates that representative samples used to generate grimoires can indeed effectively affect the final performance of grimoires; on the other hand, it indicates that the optimal grimoire for different tasks is not the same. That is, we can optimize the grimoire at the task level and select the grimoire auxiliary weak language model for different tasks. \textsc{SleIcl} proposed by us is further based on this, that is, grimoire optimization at the sample level. However, although the similarity-based method and classifier-based method we have implemented have some improvement compared to the baseline, they still cannot comprehensively surpass single grimoire. The possible reason is that similarity-based method is limited to filtering using semantic similarity, while the neural network structure constructed by the classifier-based method is still relatively simple, and its training data is currently limited.

At the task dataset level, in sentiment analysis and topic classification, the grimoire method (single grimoire and \textsc{SleIcl}) has a small performance improvement over the baseline. In natural language inference and hate speech detection, the performance of grimoire method improved significantly and was more stable compared with the baseline, indicating that the current capacity of weak language model is sufficient for relatively simple classification tasks. For more difficult tasks (such as natural language inference requiring deep semantics and simple reasoning), grimoire method can effectively make up for the shortcomings of weak language model.

In addition, we found that four PGs outperformed their corresponding SG and two SGs outperformed their corresponding PG in 12 categories of single grimoire, suggesting that GPT3.5-Turbo may favor more complex and detailed grimoire. However, in other small models tested (see appendix for detailed results), we do not find that small models are significantly inclined to SG, which indicates that we cannot simply select the best grimoire for a language model, and further demonstrates the necessity of exploring grimoire ranking methods.

\begin{table*}[!t]
    \centering
    \resizebox{\linewidth}{!}{
    \begin{threeparttable}
        \caption{Prediction accuracy difference of grimoire method relative to baseline}
        \label{tab:Difference}
        \begin{tabular}{ccccccccccc}
            \toprule
            \multirow{2}{*}[-0.8ex]{DIFF} & \multirow{2}{*}[-0.8ex]{LLM} & \multicolumn{2}{c}{Sentiment Analysis} & \multicolumn{2}{c}{Topic Classification} & \multicolumn{2}{c}{\makecell[l]{Natural Language \\ Inference}} & \multicolumn{2}{c}{Hate Speech Detection} & \multirow{2}{*}[-0.8ex]{AVG} \\
            \cmidrule(lr){3-4} \cmidrule(lr){5-6} \cmidrule(lr){7-8} \cmidrule(lr){9-10}
             & & SST5 & Subj & AgNews & TREC & RTE & QNLI & hate\_sp18 & ethos & \\
            \midrule
            \multirow{5}{*}{\makecell[c]{Max(Single Grimoire) \\ \& \\ Zero-Shot}} & GPT-3.5-Turbo & \cellcolor[HTML]{CDFFCD}+1.40\% & \cellcolor[HTML]{3E6B79}\textcolor{white}{+26.00\%} & \cellcolor[HTML]{CDFFCD}+1.49\% & \cellcolor[HTML]{CDFFCD}+1.32\% & \cellcolor[HTML]{CDFFCD}+3.80\% & \cellcolor[HTML]{528F95}+12.00\% & \cellcolor[HTML]{99DABE}+7.51\% & \cellcolor[HTML]{CDFFCD}+4.13\% & \cellcolor[HTML]{99DABE}+7.21\% \\
            & LLaMA2-70B-Chat & \cellcolor[HTML]{F6DBDD}-1.03\% & \cellcolor[HTML]{528F95}+16.16\% & \cellcolor[HTML]{99DABE}+5.69\% & \cellcolor[HTML]{3E6B79}\textcolor{white}{+20.11\%} & \cellcolor[HTML]{CDFFCD}+1.60\% & \cellcolor[HTML]{528F95}+14.53\% & \cellcolor[HTML]{99DABE}+9.67\% & \cellcolor[HTML]{528F95}+11.40\% & \cellcolor[HTML]{99DABE}+9.77\% \\
            & LLaMA2-13B-Chat & \cellcolor[HTML]{CDFFCD}+3.29\% & \cellcolor[HTML]{2F4858}\textcolor{white}{+32.34\%} & \cellcolor[HTML]{528F95}+19.06\% & \cellcolor[HTML]{528F95}+16.82\% & \cellcolor[HTML]{99DABE}+9.95\% & \cellcolor[HTML]{528F95}+10.44\% & \cellcolor[HTML]{2F4858}\textcolor{white}{+40.91\%} & \cellcolor[HTML]{528F95}+14.68\% & \cellcolor[HTML]{528F95}+18.44\% \\
            & Baichuan2-7B-Chat & \cellcolor[HTML]{528F95}+10.82\% & \cellcolor[HTML]{3E6B79}\textcolor{white}{+28.66\%} & \cellcolor[HTML]{2F4858}\textcolor{white}{+37.88\%} & \cellcolor[HTML]{2F4858}\textcolor{white}{+52.23\%} & \cellcolor[HTML]{528F95}+11.40\% & \cellcolor[HTML]{99DABE}+7.01\% & \cellcolor[HTML]{2F4858}\textcolor{white}{+63.61\%} & \cellcolor[HTML]{3E6B79}\textcolor{white}{+29.30\%} & \cellcolor[HTML]{2F4858}\textcolor{white}{+30.11\%} \\
            & Phi-2 & \cellcolor[HTML]{99DABE}+5.34\% & \cellcolor[HTML]{2F4858}\textcolor{white}{+37.04\%} & \cellcolor[HTML]{2F4858}\textcolor{white}{+32.51\%} & NaN & \cellcolor[HTML]{99DABE}+9.76\% & NaN & \cellcolor[HTML]{2F4858}\textcolor{white}{+39.15\%} & \cellcolor[HTML]{528F95}+16.05\% & \cellcolor[HTML]{3E6B79}\textcolor{white}{+23.31\%} \\
            \specialrule{0em}{1pt}{1pt}
            \multirow{4}{*}{\makecell[c]{Max(Single Grimoire) \\ \& \\ Few-Shot}} & GPT-3.5-Turbo & \cellcolor[HTML]{CDFFCD}+1.20\% & \cellcolor[HTML]{3E6B79}\textcolor{white}{+21.33\%} & \cellcolor[HTML]{F6DBDD}-0.44\% & \cellcolor[HTML]{99DABE}+5.25\% & \cellcolor[HTML]{CDFFCD}+2.40\% & \cellcolor[HTML]{CDFFCD}+4.60\% & \cellcolor[HTML]{528F95}+17.96\% & \cellcolor[HTML]{99DABE}+8.47\% & \cellcolor[HTML]{99DABE}+7.60\% \\
            & LLaMA2-70B-Chat & \cellcolor[HTML]{CDFFCD}+1.40\% & \cellcolor[HTML]{CDFFCD}+2.05\% & \cellcolor[HTML]{99DABE}+5.20\% & \cellcolor[HTML]{528F95}+16.20\% & \cellcolor[HTML]{CDFFCD}+0.20\% & \cellcolor[HTML]{528F95}+10.13\% & \cellcolor[HTML]{3E6B79}\textcolor{white}{+25.74\%} & \cellcolor[HTML]{3E6B79}\textcolor{white}{+25.32\%} & \cellcolor[HTML]{528F95}+10.78\% \\
            & LLaMA2-13B-Chat & \cellcolor[HTML]{FFD2E5}-5.60\% & \cellcolor[HTML]{528F95}+17.52\% & \cellcolor[HTML]{3E6B79}\textcolor{white}{+28.75\%} & \cellcolor[HTML]{528F95}+11.57\% & \cellcolor[HTML]{99DABE}+6.83\% & \cellcolor[HTML]{99DABE}+5.58\% & \cellcolor[HTML]{528F95}+10.96\% & \cellcolor[HTML]{CDFFCD}+2.40\% & \cellcolor[HTML]{99DABE}+9.75\% \\
            & Baichuan2-7B-Chat & \cellcolor[HTML]{528F95}+14.87\% & \cellcolor[HTML]{CDFFCD}+1.24\% & \cellcolor[HTML]{2F4858}\textcolor{white}{+39.50\%} & \cellcolor[HTML]{2F4858}\textcolor{white}{+30.39\%} & \cellcolor[HTML]{CDFFCD}+4.87\% & \cellcolor[HTML]{CDFFCD}+4.33\% & \cellcolor[HTML]{528F95}+14.78\% & \cellcolor[HTML]{528F95}+11.30\% & \cellcolor[HTML]{528F95}+15.16\% \\
            & Phi-2 & \cellcolor[HTML]{F6DBDD}-1.04\% & \cellcolor[HTML]{99DABE}+8.57\% & NaN & \cellcolor[HTML]{528F95}+19.22\% & NaN & NaN & \cellcolor[HTML]{528F95}+10.08\% & \cellcolor[HTML]{F6DBDD}-1.16\% & \cellcolor[HTML]{99DABE}+7.13\% \\
            \specialrule{0em}{0.5pt}{0.5pt}
            \midrule[0.1pt]
            \specialrule{0em}{0.5pt}{0.5pt}
            \multirow{4}{*}{\makecell[c]{\textsc{SleIcl} (Classifier-based) \\ \& \\ Zero-Shot}} & GPT-3.5-Turbo & \cellcolor[HTML]{F6DBDD}-1.63\% & \cellcolor[HTML]{3E6B79}\textcolor{white}{+20.13\%} & \cellcolor[HTML]{F6DBDD}-0.77\% & \cellcolor[HTML]{F6DBDD}-2.66\% & \cellcolor[HTML]{CDFFCD}+1.40\% & \cellcolor[HTML]{99DABE}+6.92\% & \cellcolor[HTML]{CDFFCD}+1.17\% & \cellcolor[HTML]{CDFFCD}+2.60\% & \cellcolor[HTML]{CDFFCD}+3.39\% \\
            & LLaMA2-70B-Chat & \cellcolor[HTML]{F6DBDD}-3.33\% & \cellcolor[HTML]{99DABE}+7.10\% & \cellcolor[HTML]{F6DBDD}-0.11\% & \cellcolor[HTML]{99DABE}+6.00\% & \cellcolor[HTML]{F6DBDD}-2.33\% & \cellcolor[HTML]{99DABE}+6.33\% & \cellcolor[HTML]{CDFFCD}+1.99\% & \cellcolor[HTML]{99DABE}+8.52\% & \cellcolor[HTML]{CDFFCD}+3.02\% \\
            & LLaMA2-13B-Chat & \cellcolor[HTML]{528F95}+12.64\% & \cellcolor[HTML]{CDFFCD}+1.88\% & \cellcolor[HTML]{528F95}+13.66\% & \cellcolor[HTML]{3E6B79}\textcolor{white}{+24.07\%} & \cellcolor[HTML]{FFD2E5}-8.94\% & \cellcolor[HTML]{FFD2E5}-5.72\% & \cellcolor[HTML]{2F4858}\textcolor{white}{+34.35\%} & \cellcolor[HTML]{F6DBDD}-1.69\% & \cellcolor[HTML]{99DABE}+8.78\% \\
            & Baichuan2-7B-Chat & \cellcolor[HTML]{CDFFCD}+3.58\% & \cellcolor[HTML]{2F4858}\textcolor{white}{+42.80\%} & \cellcolor[HTML]{2F4858}\textcolor{white}{+30.82\%} & \cellcolor[HTML]{2F4858}\textcolor{white}{+37.85\%} & \cellcolor[HTML]{99DABE}+6.93\% & \cellcolor[HTML]{F6DBDD}-1.11\% & \cellcolor[HTML]{2F4858}\textcolor{white}{+50.18\%} & \cellcolor[HTML]{3E6B79}\textcolor{white}{+25.42\%} & \cellcolor[HTML]{3E6B79}\textcolor{white}{+24.56\%} \\
            & Phi-2 & \cellcolor[HTML]{CDFFCD}+3.19\% & \cellcolor[HTML]{2F4858}\textcolor{white}{+42.89\%} & \cellcolor[HTML]{3E6B79}\textcolor{white}{+24.34\%} & NaN & NaN & NaN & \cellcolor[HTML]{2F4858}\textcolor{white}{+32.85\%} & \cellcolor[HTML]{528F95}+13.26\% & \cellcolor[HTML]{3E6B79}\textcolor{white}{+23.31\%} \\
            \specialrule{0em}{1pt}{1pt}
            \multirow{4}{*}{\makecell[c]{\textsc{SleIcl} (Classifier-based) \\ \& \\ Few-Shot}} & GPT-3.5-Turbo & \cellcolor[HTML]{F6DBDD}-1.83\% & \cellcolor[HTML]{528F95}+15.47\% & \cellcolor[HTML]{F6DBDD}-2.70\% & \cellcolor[HTML]{CDFFCD}+1.27\% & 0.00\% & \cellcolor[HTML]{F6DBDD}-0.48\% & \cellcolor[HTML]{528F95}+11.62\% & \cellcolor[HTML]{99DABE}+6.93\% & \cellcolor[HTML]{CDFFCD}+3.78\% \\
            & LLaMA2-70B-Chat & \cellcolor[HTML]{F6DBDD}-0.90\% & \cellcolor[HTML]{FFD2E5}-7.01\% & \cellcolor[HTML]{F6DBDD}-0.60\% & \cellcolor[HTML]{CDFFCD}+2.09\% & \cellcolor[HTML]{F6DBDD}-3.73\% & \cellcolor[HTML]{CDFFCD}+1.93\% & \cellcolor[HTML]{528F95}+18.06\% & \cellcolor[HTML]{3E6B79}\textcolor{white}{+22.44\%} & \cellcolor[HTML]{CDFFCD}+4.04\% \\
            & LLaMA2-13B-Chat & \cellcolor[HTML]{CDFFCD}+3.75\% & \cellcolor[HTML]{FFCBFD}-12.94\% & \cellcolor[HTML]{3E6B79}\textcolor{white}{+23.35\%} & \cellcolor[HTML]{528F95}+18.82\% & \cellcolor[HTML]{FFCBFD}-12.06\% & \cellcolor[HTML]{FFCBFD}-10.58\% & \cellcolor[HTML]{CDFFCD}+4.40\% & \cellcolor[HTML]{FFCBFD}-13.97\% & \cellcolor[HTML]{CDFFCD}+0.10\% \\
            & Baichuan2-7B-Chat & \cellcolor[HTML]{99DABE}+7.63\% & \cellcolor[HTML]{528F95}+15.38\% & \cellcolor[HTML]{2F4858}\textcolor{white}{+32.44\%} & \cellcolor[HTML]{528F95}+16.01\% & \cellcolor[HTML]{CDFFCD}+0.40\% & \cellcolor[HTML]{F6DBDD}-3.79\% & \cellcolor[HTML]{CDFFCD}+1.35\% & \cellcolor[HTML]{99DABE}+7.42\% & \cellcolor[HTML]{99DABE}+9.61\% \\
            & Phi-2 & \cellcolor[HTML]{F6DBDD}-3.19\% & \cellcolor[HTML]{528F95}+14.42\% & NaN & \cellcolor[HTML]{99DABE}+6.08\% & NaN & NaN & \cellcolor[HTML]{CDFFCD}+3.78\% & \cellcolor[HTML]{F6DBDD}-3.95\% & \cellcolor[HTML]{CDFFCD}+3.43\% \\
            \bottomrule
        \end{tabular}
        \begin{tablenotes}
            \footnotesize
            \item \textit{Note}: Max(Single Grimoire) indicates the best performance among all Single Grimoire methods; Positive differences will be highlighted in green, with darker colors being greater: \colorbox[HTML]{CDFFCD}{\textless 5\%}, \colorbox[HTML]{99DABE}{5\% $\sim$ 10\%}, \colorbox[HTML]{528F95}{10\% $\sim$ 20\%}, \colorbox[HTML]{3E6B79}{\textcolor{white}{20\% $\sim$ 30\%}}, \colorbox[HTML]{2F4858}{\textcolor{white}{\textgreater 30\%}}; The negative difference will be highlighted in red, and the smaller the difference, the darker the color: \colorbox[HTML]{F6DBDD}{\textgreater -5\%}, \colorbox[HTML]{FFD2E5}{-5\% $\sim$ -10\%}, \colorbox[HTML]{FFCBFD}{\textless -10\%}. NaN indicates that the number of valid experimental data is too small to give a reliable accuracy rate.
        \end{tablenotes} 
    \end{threeparttable}
    }
\end{table*}

In Table \ref{tab:Difference}, we provide a detailed presentation of the differences between the best performance of all single grimoires (Max(single grimoire)) and baseline performance on each task dataset, as well as the differences between \textsc{SleIcl} (classifier-based) and baseline performance. Among all language models, the Max(single grimoire) almost comprehensively exceeds baseline, and it can be found that the smaller the language model, the greater the performance improvement compared with baseline under the support of grimoire. For example, Baichuan2-7B has improved by 30.11\% and 15.16\% relative to zero-shot and few-shot, respectively, while GPT3.5-Turbo has improved by 7.21\% and 7.60\% relative to zero-shot and few-shot, respectively. On the other hand, this also indicates that for weak language model, the grimoire method can significantly outperform the performance of few-shot (n=4) and has a wide range of applicability (weak language model from 175B to 7B have significant improvements). At the same time, we can once again observe that on all weak language models, the grimoire method performs better in natural language interference and hate speech detection tasks, which further demonstrates that the grimoire method can effectively enhance the capability of weak language models. 

\begin{figure}[!t]
    \centering
    \includegraphics[width=\linewidth]{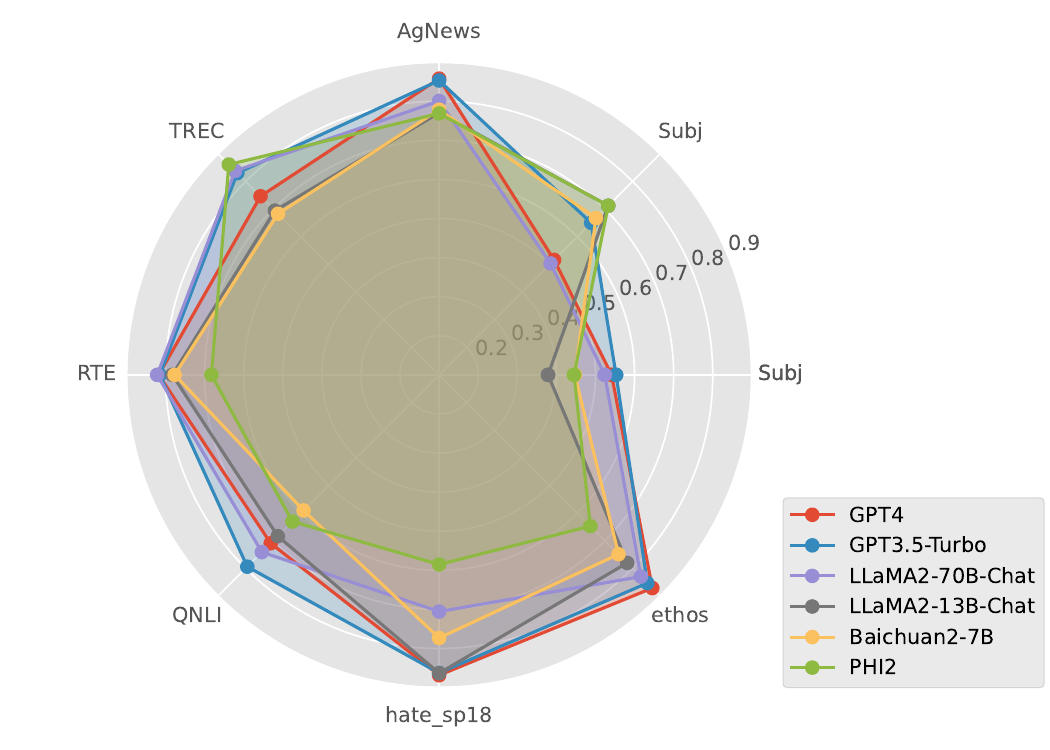}
    \caption{Radar Chart comparing GPT-4 results in zero-shot prompting with other models' results in Max(Single Grimoire) setting.}
    \label{fig:radar}
\end{figure}

In all weak language models, although the average performance of the \textsc{SleIcl}(classifier-based) has improved compared to the baseline, it still cannot surpass the baseline on some datasets. This indicates that although the \textsc{SleIcl}(classifier-based) has improved compared to the \textsc{SleIcl}(similarity-based), it still has not achieved the ideal performance. In addition, we can still observe the pattern that smaller language models can achieve more significant improvements. For example, on almost all datasets, Baichuan2-7B has a significant improvement compared to baseline in the \textsc{SleIcl}(classifier-based), while other larger language models do not perform so well. Furthermore, we can observe from Figure~\ref{fig:radar} that with the use of grimoires, weak language models have the potential to surpass GPT4-1106-preview under zero-shot settings. Even on some datasets, the performance of weak language models exceeds that of larger-scale language models. For instance, on the TREC and Subj datasets, the PHI2 model with only 2.7 billion parameters outperforms GPT4-1106-preview.

\section{Conclusion}

In this paper, we introduce a method named \textsc{SleIcl}, predicated on utilizing strong language models to learn from representative samples and distill skills for solving specific tasks, thereby enhancing the proficiency of weak language models in these tasks. The synthesized skills within this framework are termed \textbf{grimoire}. To diversify the grimoire categories and comprehensively examine their impacts, we developed four distinct representative sample selection methods (KCS, HCS, HSS, RSS) and a zero-shot approach, alongside two grimoire generation templates (PG and SG), culminating in the creation of 10 types of single grimoires. Building on this, we formulated a grimoire ranking method, aimed at automating the selection of the most suitable grimoire for various models and tasks at the sample level. Ultimately, we evaluated 5 models across 8 datasets under 4 task types, demonstrating that \textsc{SleIcl} can substantially enhance the performance of weak language models with varying parameter sizes on diverse tasks, with smaller models exhibiting more pronounced improvements. Remarkably, on certain datasets, weak language models, with the aid of our method, outperformed GPT4-1106-preview in zero-shot scenarios. However, while our grimoire ranking method showed some improvements over Zero-shot and Few-shot approaches, it did not surpass the performance of the best single grimoire results, suggesting that the classifier-based method has potential for further optimization. Additionally, the representative sample selection method presents an avenue for further exploration to expand the variety of grimoires available for weak language models across diverse tasks.

\newpage






\bibliographystyle{named}
\bibliography{ijcai24}

\newpage
\appendix
\section*{Appendix}

\section{Datasets and models}
Some additional details about all the datasets used in this paper are shown in Table \ref{tab:dataset}, including the number of dataset sample labels (\#Class), the number of training set samples, and the number of test set samples.

\begin{table}[htp]
    \centering
    \resizebox{\linewidth}{!}{
    \begin{threeparttable}
        \caption{Dataset information}
        \label{tab:dataset}
        \begin{tabular}{llccc}
            \toprule
            Task & Dataset & \#Class & \#Train  & \#Eval        \\
            \midrule
            \multirow{2}{*}{Sentiment Analysis} & SST5 & 5 & 8544 & 1101 \\
             & Subj & 2 & 10000 & - \\
            \midrule[0.5pt]
            \multirow{2}{*}{Topic Classification} & AgNews & 4 & 120000 & 7600 \\
             & TREC & 6 & 5452 & 500 \\
            \midrule[0.5pt]
            \multirow{2}{*}{\makecell[l]{Natural Language \\ Inference}} & RTE & 2 & 2490 & 277 \\
             & QNLI & 2 & 104743 & 5463 \\
            \midrule[0.5pt]
            \multirow{2}{*}{\makecell[l]{Hate Speech \\ Detection}} & hate\_sp18 & 2 & 10944 & - \\
             & ethos & 2 & 998 & - \\
            \bottomrule
        \end{tabular}
        \begin{tablenotes}
            \footnotesize
            \item \textit{Note}: - indicates that the dataset is not pre-divided into the training set and the evaluation set.
        \end{tablenotes}
    \end{threeparttable}}
\end{table}

All of the language models used in our evaluation are shown in Table \ref{tab:models}, sorted by release date.

\begin{table}[htp]
    \centering
    \resizebox{\linewidth}{!}{
    \begin{threeparttable}
        \caption{Models Sorted by Release Date}
        \label{tab:models}
        \begin{tabular}{lllll}
            \toprule
            Model & Parm. & Type & Publisher & Release  \\
            \midrule
             GPT3.5-Turbo & 175B$^*$ & Chat & OpenAI & 2023.06$^*$ \\
             LLaMA2 & 70B & Chat & Meta & 2023.07 \\
             LLaMA2 & 13B & Chat & Meta & 2023.07 \\
             Baichuan2 & 7B & Chat & Baichuan Inc. & 2023.09 \\
             GPT4-1106-preview & NaN & Chat & OpenAI & 2023.11 \\
             PHI2 & 2.7B & Base & Microsoft & 2023.12 \\
            \bottomrule
        \end{tabular}
        \begin{tablenotes}
            \footnotesize
            \item \textit{Note}: * indicates an estimated value, NaN signifies the absence of publicly available data, and 175B represents 175 billion.
        \end{tablenotes}
    \end{threeparttable}}
\end{table}

\section{Dual-tower deep neural network classifier}

\subsection{Feature engineering}

In order to train the neural network constructed in the classifier-based method, we primarily designed four categories of features to identify language models, tasks, test samples, and grimoires respectively. We selected the parameter count as the feature to identify language models, as the performance of large models is directly related to their parameter scale. For tasks, we selected three features related to task category and task description. For test samples, we chose the text length of the sample and its embedding vector as corresponding features. Regarding grimoires, we selected their type, text length, embedding vector, and representative sample selection method as features. The detailed information about the features can be found in Table \ref{tab:feature}.

In the processing of raw features, we utilize MPNet~\cite{mpnet_20_NIPS_Microsoft} to embed task descriptions, test questions, and grimoire, generating corresponding 768-dimensional dense vectors. For categorical features, we uniformly apply one-hot encoding. As for features related to text length, we initially apply data binning and subsequently employ one-hot encoding based on this binned data.

\begin{table*}[htp]
 \caption{Features of the classifier}
  \centering
  \resizebox{\linewidth}{!}{
  \begin{tabular}{lllcc}
    \toprule
    Feature type & Feature & Description & Original data type & Embedding method \\
    \midrule
    LLM-based & llm\_param\_cnt\_type & LLM parameter scale type & Enum & One-hot \\
    \midrule
    \multirow{3}{*}{Task-based} & task\_type & Task type & Enum & One-hot \\
     & task\_desc\_len\_type & Types of the task description length & Int & One-hot \\
     & task\_desc\_emb & Embedding of the task description & Str & Dense \\
     \midrule
    \multirow{2}{*}{Question-based} & question\_len\_type & Types of question text length & Int & One-hot \\
     & question\_emb & Embedding of the question & Str & Dense \\
     \midrule
    \multirow{4}{*}{Grimoire-based} & grimoire\_type & Grimoire type & Enum & One-hot \\
     & grimoire\_len\_type & Types of grimoire length & Int & One-hot \\
     & grimoire\_sample\_method & Sample selection methods of grimoire & Enum & One-hot \\
     & grimoire\_emb & Embedding of the grimoire & Str & Dense\\
    \bottomrule
  \end{tabular}}
  \label{tab:feature}
\end{table*}

\subsection{Classifier architecture}

\begin{figure}[t!]
    \centering
    \includegraphics[width=0.8\linewidth]{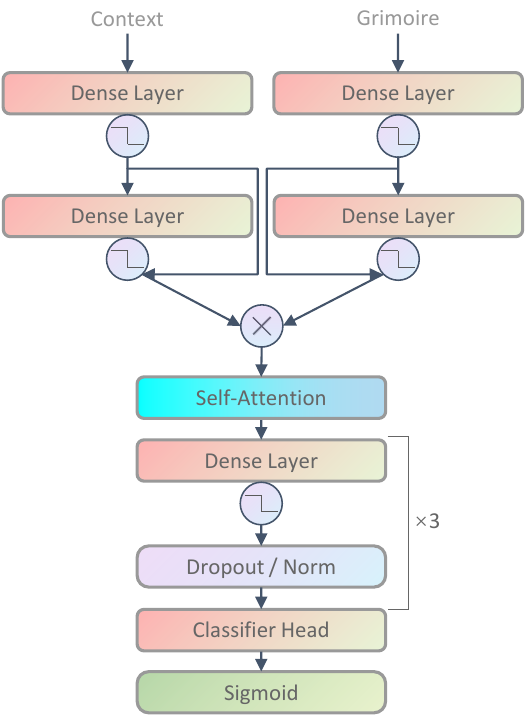}
    \caption{Architecture of the classifier. Within the three similar forward propagation modules following self-attention, the first two employ dropouts, while the final one employs normalization.}
    \label{fig:archi}
\end{figure}

After obtaining vector representations for individual features, we concatenate all vectors from LLM-based, Task-based, and Question-based sources to form an enriched context vector, representing an augmented user query. Simultaneously, the concatenated feature vectors from the Grimoire-based source form the grimoire vector. Subsequently, modeling the relationship between the context and grimoire to derive utility values constitutes the objective of the classifier, as illustrated in Figure~\ref{fig:archi}. This architecture consists of two independent branches, with each branch processing context and grimoire separately through two fully connected layers. Additionally, each branch incorporates residual block structures. Following this, the processed vectors are stacked, and a joint representation is formed using the self-attention mechanism. After three similar forward propagation modules, the resulting representation passes through a classifier head and, via a sigmoid activation, yields the prediction values.

\subsection{Training process}

The training dataset for the classifier is derived from intermediate output results obtained from experiments with similarity-based ranking approach, encompassing over 30,000+ data points. During training, a batch size of 1024 is employed, with a learning rate set to 0.001. The loss function utilized is binary cross-entropy, and the Adam optimizer is employed. We consider the model weights at 500 epochs as fixed for subsequent experiments. The training of this model is conducted using a single NVIDIA A100 40GB GPU.

\section{Detailed evaluation results of other language models}
Detailed evaluation results of LLaMA2-70B-Chat, LLaMA2-13B-Chat, Baichuan2-7B-Chat and Phi-2 are supplemented in Table \ref{tab:Detailed evaluation results of LLaMA2-70B-Chat}-\ref{tab:Detailed evaluation results of Phi-2}. Due to the poor instruction following ability of Phi-2, there are less than 500 valid experimental data in multiple evaluation items, which cannot guarantee the reliability of accuracy rate, so some evaluation results are missing.

Overall, the best performance on various datasets is mostly achieved by the grimoire method, which aligns with our previously drawn conclusion that grimoire can effectively enhance the performance of weak language models across different tasks. Secondly, for these four language models, the best performance on individual datasets is not concentrated on a specific grimoire method, reaffirming that it is not possible to simply identify a universally optimal grimoire for all tasks and language models. Therefore, the grimoire ranking method is necessary. Furthermore, although on Baichuan2-7B-Chat and Phi-2, the average performance of \textsc{SleIcl}(classifier-based) is the best, \textsc{SleIcl} still cannot consistently outperform all single grimoire methods on all weak language models. Therefore, further optimization of the grimoire ranking method is still needed in the future. Finally, we still observe the pattern that smaller language models tend to exhibit a higher improvement over the baseline under the grimoire method. Additionally, we find that the average performance of LLaMA2-13B-Chat on RSS-PG significantly surpasses that of LLaMA2-70B-Chat in the same series but with a larger parameter size, indicating that the benefits of the grimoire method for weak language models are quite significant.

\

\begin{table*}[!t]
    \centering
    \resizebox{\linewidth}{!}{
    \begin{threeparttable}
        \caption{Detailed evaluation results of LLaMA2-70B-Chat enhanced by grimoire.}
        \label{tab:Detailed evaluation results of LLaMA2-70B-Chat}
        \begin{tabular}{cccccccccccc}
            \toprule
            \multirow{2}{*}[-0.8ex]{Model} & \multirow{2}{*}[-0.8ex]{Submodel} & \multirow{2}{*}[-0.8ex]{n/k-shot} & \multicolumn{2}{c}{Sentiment analysis} & \multicolumn{2}{c}{Topic classification} & \multicolumn{2}{c}{\makecell[l]{Natural language \\ inference}} & \multicolumn{2}{c}{Hate speech detection} & \multirow{2}{*}[-0.8ex]{Avg} \\
           \cmidrule(lr){4-5} \cmidrule(lr){6-7} \cmidrule(lr){8-9} \cmidrule(lr){10-11} 
            & & & SST5 & Subj & AgNews & TREC & RTE & QNLI & hate\_sp18 & ethos & \\
            \midrule
            \multirow{2}{*}{Baseline} & Zero-Shot & - & \textbf{53.40\%} & 34.15\% & 74.38\% & 63.69\% & 80.53\% & 59.60\% & 60.88\% & 71.59\% & 62.28\% \\
            & Few-Shot & n=4 & 50.97\% & \underline{48.26\%} & 74.87\% & 67.60\% & \underline{81.93\%} & 64.00\% & 44.81\% & 57.67\% & 61.26\% \\
            \specialrule{0em}{1pt}{1pt}
            \midrule
            \multirow{12}{*}{Single Grimoire} & KCS-PG & k=4 & 52.20\% & \textbf{50.31\%} & \underline{77.52\%} & 68.45\% & 73.00\% & 58.93\% & 55.23\% & \textbf{82.99\%} & 64.83\% \\
            & KCS-SG & k=4 & \underline{52.37\%} & 36.72\% & 75.12\% & 67.28\% & 81.73\% & 64.47\% & 55.77\% & 73.16\% & 63.33\% \\
            & HCS-PG & k=4 & 50.71\% & 37.38\% & 73.68\% & 73.80\% & 77.67\% & 65.80\% & 60.51\% & 73.27\% & 64.10\% \\
            & HCS-SG & k=4 & 48.20\% & 37.35\% & 74.72\% & 74.88\% & 79.27\% & 64.93\% & \underline{68.35\%} & 73.24\% & 65.12\% \\
            & HSS-PG & k=4 (r=1.0) & 49.55\% & 39.67\% & 76.80\% & 60.81\% & \textbf{82.13\%} & 58.20\% & \textbf{70.55\%} & \underline{80.82\%} & 64.82\% \\
            & HSS-SG & k=4 (r=1.0) & 49.39\% & 39.07\% & 74.93\% & 57.39\% & 81.80\% & 66.80\% & 61.95\% & 71.66\% & 62.87\% \\
            & HSS-PG & k=4 (r=0.5) & 47.95\% & 42.90\% & \textbf{80.07\%} & \textbf{83.80\%} & 76.80\% & \underline{66.93\%} & 66.51\% & 79.03\% & \textbf{68.00\%} \\
            & HSS-SG & k=4 (r=0.5) & 49.59\% & 36.44\% & 73.33\% & \underline{78.03\%} & 79.07\% & 63.47\% & 55.00\% & 75.39\% & 63.79\% \\
            & RSS-PG & k=4 & 50.53\% & 37.24\% & 71.36\% & 66.40\% & 79.13\% & \textbf{74.13\%} & 64.66\% & 77.81\% & 65.16\% \\
            & RSS-SG & k=4 & 46.51\% & 41.36\% & 69.49\% & 69.33\% & 80.60\% & 63.20\% & 57.66\% & 68.79\% & 62.12\% \\
            & Zero-Shot-PG & - & 46.86\% & 43.32\% & 76.40\% & 60.49\% & 75.73\% & 58.53\% & 65.81\% & 76.35\% & 62.94\% \\
            & Zero-Shot-SG & - & 49.52\% & 34.99\% & 73.27\% & 52.53\% & 79.80\% & 55.80\% & 62.90\% & 74.55\% & 60.42\% \\
            \specialrule{0em}{1pt}{1pt}
            \midrule
            \multirow{2}{*}{\textsc{SleIcl}} & Similarity-based & - & 50.90\% & 39.93\% & 74.67\% & 74.37\% & 79.33\% & 64.27\% & 61.69\% & 70.30\% & 64.43\% \\
            & Classifier-based & - & 50.07\% & 41.25\% & 74.27\% & 69.69\% & 78.20\% & 65.93\% & 62.87\% & 80.11\% & \underline{65.30\%} \\
            \bottomrule
        \end{tabular}
        \begin{tablenotes}
            \footnotesize
            \item \textit{Note}: - indicates that this hyper-parameter is invalid for the current test; n-shot indicates that n samples will be provided for each prediction; k-shot provides a selection of k samples under each label to generate grimoire; r represents the sampling ratio of hard samples. The best performance in each column will be bolded, and the second-best performance will be underlined.
        \end{tablenotes}
    \end{threeparttable}
    }
\end{table*}

\begin{table*}[!t]
    \centering
    \resizebox{\linewidth}{!}{
    \begin{threeparttable}
        \caption{Detailed evaluation results of LLaMA2-13B-Chat enhanced by grimoire.}
        \label{tab:Detailed evaluation results of LLaMA2-13B-Chat}
        \begin{tabular}{cccccccccccc}
            \toprule
            \multirow{2}{*}[-0.8ex]{Model} & \multirow{2}{*}[-0.8ex]{Submodel} & \multirow{2}{*}[-0.8ex]{n/k-shot} & \multicolumn{2}{c}{Sentiment analysis} & \multicolumn{2}{c}{Topic classification} & \multicolumn{2}{c}{\makecell[l]{Natural language \\ inference}} & \multicolumn{2}{c}{Hate speech detection} & \multirow{2}{*}[-0.8ex]{Avg} \\
           \cmidrule(lr){4-5} \cmidrule(lr){6-7} \cmidrule(lr){8-9} \cmidrule(lr){10-11} 
            & & & SST5 & Subj & AgNews & TREC & RTE & QNLI & hate\_sp18 & ethos & \\
            \midrule
            \multirow{2}{*}{Baseline} & Zero-Shot & - & 34.57\% & 38.85\% & 57.98\% & 52.60\% & 68.31\% & 57.92\% & 45.40\% & 63.36\% & 52.37\% \\
            & Few-Shot & n=4 & \underline{43.46\%} & 53.67\% & 48.29\% & 57.85\% & 71.43\% & 62.78\% & 75.35\% & 75.64\% & 61.06\% \\
            \specialrule{0em}{1pt}{1pt}
            \midrule
            \multirow{12}{*}{Single Grimoire} & KCS-PG & k=4 & 35.53\% & 69.19\% & \underline{75.33\%} & 69.02\% & 48.07\% & 64.15\% & 72.67\% & 76.06\% & 63.75\% \\
            & KCS-SG & k=4 & 36.79\% & 47.37\% & 74.62\% & 51.88\% & 69.39\% & 60.40\% & \underline{83.40\%} & 68.30\% & 61.52\% \\
            & HCS-PG & k=4 & 35.14\% & 64.88\% & 74.13\% & 62.20\% & 56.07\% & 59.06\% & 80.52\% & 76.55\% & 63.57\% \\
            & HCS-SG & k=4 & 35.53\% & 40.88\% & 72.18\% & 49.10\% & 63.80\% & 64.53\% & \textbf{86.31\%} & \textbf{78.04\%} & 61.30\% \\
            & HSS-PG & k=4 (r=1.0) & 37.86\% & \textbf{71.19\%} & 73.73\% & \underline{69.42\%} & 59.07\% & \underline{67.20\%} & 56.31\% & 72.05\% & 63.35\% \\
            & HSS-SG & k=4 (r=1.0) & 34.85\% & 47.20\% & 71.91\% & 43.73\% & \textbf{78.26\%} & 58.60\% & 55.15\% & 60.85\% & 56.32\% \\
            & HSS-PG & k=4 (r=0.5) & 36.26\% & 67.91\% & 74.25\% & 51.97\% & \underline{73.27\%} & \textbf{68.36\%} & 72.95\% & 74.68\% & \underline{64.96\%} \\
            & HSS-SG & k=4 (r=0.5) & 36.17\% & 38.53\% & 34.47\% & 42.76\% & 71.27\% & 63.00\% & 60.40\% & 75.19\% & 52.72\% \\
            & RSS-PG & k=4 & 32.97\% & 69.53\% & \textbf{77.04\%} & 62.68\% & 68.13\% & 66.69\% & 70.86\% & 75.13\% & \textbf{65.38\%} \\
            & RSS-SG & k=4 & 35.46\% & 53.41\% & 61.32\% & 51.75\% & 72.55\% & 62.13\% & 34.96\% & 67.00\% & 54.82\% \\
            & Zero-Shot-PG & - & 35.01\% & \underline{69.79\%} & 52.00\% & 46.72\% & 52.87\% & 60.75\% & 78.16\% & 70.21\% & 58.19\% \\
            & Zero-Shot-SG & - & 34.96\% & 53.06\% & 49.83\% & 47.34\% & 59.89\% & 57.37\% & 76.47\% & \underline{77.54\%} & 57.06\% \\
            \specialrule{0em}{1pt}{1pt}
            \midrule
            \multirow{2}{*}{\textsc{SleIcl}} & Similarity-based & - & 33.74\% & 50.33\% & 66.38\% & 52.58\% & 58.49\% & 62.60\% & 70.56\% & 73.05\% & 58.47\% \\
            & Classifier-based & - & \textbf{47.21\%} & 40.73\% & 71.64\% & \textbf{76.67\%} & 59.37\% & 52.20\% & 79.75\% & 61.67\% & 61.16\%\\
            \bottomrule
        \end{tabular}
        \begin{tablenotes}
            \footnotesize
            \item \textit{Note}: - indicates that this hyper-parameter is invalid for the current test; n-shot indicates that n samples will be provided for each prediction; k-shot provides a selection of k samples under each label to generate grimoire; r represents the sampling ratio of hard samples. The best performance in each column will be bolded, and the second-best performance will be underlined.
        \end{tablenotes}
    \end{threeparttable}
    }
\end{table*}

\begin{table*}[!t]
    \centering
    \resizebox{\linewidth}{!}{
    \begin{threeparttable}
        \caption{Detailed evaluation results of Baichuan2-7B-Chat enhanced by grimoire.}
        \label{tab:Detailed evaluation results of Baichuan2-7B-Chat}
        \begin{tabular}{cccccccccccc}
            \toprule
            \multirow{2}{*}[-0.8ex]{Model} & \multirow{2}{*}[-0.8ex]{Submodel} & \multirow{2}{*}[-0.8ex]{n/k-shot} & \multicolumn{2}{c}{Sentiment analysis} & \multicolumn{2}{c}{Topic classification} & \multicolumn{2}{c}{\makecell[l]{Natural language \\ inference}} & \multicolumn{2}{c}{Hate speech detection} & \multirow{2}{*}[-0.8ex]{Avg} \\
           \cmidrule(lr){4-5} \cmidrule(lr){6-7} \cmidrule(lr){8-9} \cmidrule(lr){10-11} 
            & & & SST5 & Subj & AgNews & TREC & RTE & QNLI & hate\_sp18 & ethos & \\
            \midrule
            \multirow{2}{*}{Baseline} & Zero-Shot & - & 33.91\% & 38.07\% & 39.96\% & 16.05\% & 66.27\% & 52.04\% & 13.70\% & 45.60\% & 38.20\% \\
            & Few-Shot & n=4 & 29.86\% & 65.49\% & 38.34\% & 37.89\% & 72.80\% & 54.72\% & 62.53\% & 63.60\% & 53.15\% \\
            \specialrule{0em}{1pt}{1pt}
            \midrule
            \multirow{12}{*}{Single Grimoire} & KCS-PG & k=4 & 39.56\% & 66.07\% & 60.57\% & 54.78\% & \textbf{77.67\%} & 57.04\% & 46.03\% & 68.21\% & 58.74\% \\
            & KCS-SG & k=4 & 39.93\% & 43.33\% & 42.32\% & 19.09\% & 72.00\% & 52.35\% & \underline{75.92\%} & 53.67\% & 49.83\% \\
            & HCS-PG & k=4 & 33.67\% & 64.87\% & 44.30\% & \underline{66.07\%} & 77.00\% & \underline{58.87\%} & 54.39\% & 68.89\% & 58.51\% \\
            & HCS-SG & k=4 & 39.07\% & 42.47\% & 44.33\% & 34.97\% & 71.40\% & \textbf{59.05\%} & \textbf{77.31\%} & 54.85\% & 52.93\% \\
            & HSS-PG & k=4 (r=1.0) & \underline{42.39\%} & 63.00\% & 36.47\% & 32.28\% & 74.33\% & 57.60\% & 50.26\% & 65.47\% & 52.73\% \\
            & HSS-SG & k=4 (r=1.0) & 42.02\% & \underline{66.73\%} & 48.99\% & 28.37\% & 70.07\% & 52.60\% & 21.54\% & 47.36\% & 47.21\% \\
            & HSS-PG & k=4 (r=0.5) & 29.30\% & 65.67\% & 70.01\% & \textbf{68.28\%} & 73.87\% & 53.47\% & 41.05\% & \textbf{74.90\%} & \underline{59.57\%} \\
            & HSS-SG & k=4 (r=0.5) & \textbf{44.73\%} & 47.07\% & \textbf{77.84\%} & 45.66\% & 69.53\% & 57.18\% & 17.60\% & 61.47\% & 52.64\% \\
            & RSS-PG & k=4 & 34.40\% & 63.40\% & 49.13\% & 48.71\% & 73.60\% & 56.09\% & 42.28\% & 67.68\% & 54.41\% \\
            & RSS-SG & k=4 & 40.32\% & 36.67\% & 45.64\% & 26.98\% & 68.80\% & 57.79\% & 15.80\% & 46.66\% & 42.33\% \\
            & Zero-Shot-PG & - & 33.22\% & 35.93\% & 39.09\% & 28.75\% & 76.13\% & 51.60\% & 49.16\% & 60.92\% & 46.85\% \\
            & Zero-Shot-SG & - & 41.13\% & 40.07\% & 57.35\% & 22.89\% & \underline{77.40\%} & 54.18\% & 28.13\% & 64.25\% & 48.18\% \\
            \specialrule{0em}{1pt}{1pt}
            \midrule
            \multirow{2}{*}{\textsc{SleIcl}} & Similarity-based & - & 36.49\% & 61.53\% & 56.84\% & 35.35\% & 77.25\% & 57.95\% & 45.12\% & 55.59\% & 53.26\% \\
            & Classifier-based & - & 37.49\% & \textbf{80.87\%} & \underline{70.78\%} & 53.90\% & 73.20\% & 50.93\% & 63.88\% & \underline{71.02\%} & \textbf{62.76\%} \\
            \bottomrule
        \end{tabular}
        \begin{tablenotes}
            \footnotesize
            \item \textit{Note}: - indicates that this hyper-parameter is invalid for the current test; n-shot indicates that n samples will be provided for each prediction; k-shot provides a selection of k samples under each label to generate grimoire; r represents the sampling ratio of hard samples. The best performance in each column will be bolded, and the second-best performance will be underlined.
        \end{tablenotes}
    \end{threeparttable}
    }
\end{table*}

\begin{table*}[!t]
    \centering
    \resizebox{\linewidth}{!}{
    \begin{threeparttable}
        \caption{Detailed evaluation results of Phi-2 enhanced by grimoire.}
        \label{tab:Detailed evaluation results of Phi-2}
        \begin{tabular}{cccccccccccc}
            \toprule
            \multirow{2}{*}[-0.8ex]{Model} & \multirow{2}{*}[-0.8ex]{Submodel} & \multirow{2}{*}[-0.8ex]{n/k-shot} & \multicolumn{2}{c}{Sentiment analysis} & \multicolumn{2}{c}{Topic classification} & \multicolumn{2}{c}{\makecell[l]{Natural language \\ inference}} & \multicolumn{2}{c}{Hate speech detection} & \multirow{2}{*}[-0.8ex]{Avg} \\
           \cmidrule(lr){4-5} \cmidrule(lr){6-7} \cmidrule(lr){8-9} \cmidrule(lr){10-11} 
            & & & SST5 & Subj & AgNews & TREC & RTE & QNLI & hate\_sp18 & ethos & \\
            \midrule
            \multirow{2}{*}{Baseline} & Zero-Shot & - & 39.18\% & 34.15\% & 44.38\% & NaN & 58.50\% & NaN & 19.37\% & 48.68\% & 40.71\% \\
            & Few-Shot & n=4 & \textbf{45.56\%} & 62.62\% & NaN & 66.88\% & NaN & NaN & 48.44\% & \textbf{65.89\%} & 57.88\% \\
            \specialrule{0em}{1pt}{1pt}
            \midrule
            \multirow{12}{*}{Single Grimoire} & KCS-PG & k=4 & 40.96\% & 39.31\% & 73.19\% & 52.78\% & NaN & NaN & 44.21\% & 62.49\% & 52.16\% \\
            & KCS-SG & k=4 & 43.67\% & 34.49\% & 53.22\% & NaN & \underline{66.71\%} & NaN & 52.19\% & 52.47\% & 50.46\% \\
            & HCS-PG & k=4 & \underline{44.52\%} & 42.00\% & 76.06\% & \textbf{86.10\%} & NaN & \textbf{63.07\%} & 36.10\% & \underline{64.73\%} & \underline{58.94\%} \\
            & HCS-SG & k=4 & 42.01\% & 41.01\% & 64.97\% & 72.33\% & \textbf{68.26\%} & \underline{61.49\%} & \textbf{58.52\%} & 60.55\% & 58.64\% \\
            & HSS-PG & k=4 (r=1.0) & 40.89\% & 52.60\% & \underline{76.38\%} & 55.37\% & NaN & NaN & 37.54\% & 56.11\% & 53.15\% \\
            & HSS-SG & k=4 (r=1.0) & 38.52\% & 37.97\% & 61.15\% & NaN & 63.92\% & 52.22\% & 24.01\% & 58.88\% & 48.10\% \\
            & HSS-PG & k=4 (r=0.5) & 41.71\% & \underline{71.19\%} & 34.40\% & 47.51\% & NaN & NaN & 29.01\% & 55.32\% & 46.52\% \\
            & HSS-SG & k=4 (r=0.5) & 42.13\% & 36.62\% & 37.16\% & 64.48\% & 61.47\% & 57.02\% & 47.03\% & 58.44\% & 50.54\% \\
            & RSS-PG & k=4 & 38.80\% & 51.99\% & \textbf{76.89\%} & 51.12\% & NaN & NaN & \underline{52.65\%} & 63.02\% & 55.75\% \\
            & RSS-SG & k=4 & 41.81\% & 36.21\% & 65.06\% & 62.13\% & NaN & NaN & 23.91\% & 48.29\% & 46.24\% \\
            & Zero-Shot-PG & - & 39.59\% & 36.46\% & 49.13\% & 52.53\% & NaN & NaN & 42.07\% & 53.21\% & 45.50\% \\
            & Zero-Shot-SG & - & 41.38\% & 33.84\% & 43.51\% & NaN & 58.51\% & NaN & 36.55\% & 59.94\% & 45.62\% \\
            \specialrule{0em}{1pt}{1pt}
            \midrule
            \multirow{2}{*}{\textsc{SleIcl}} & Similarity-based & - & 41.26\% & 50.52\% & 59.35\% & NaN & 64.96\% & NaN & 37.73\% & 55.33\% & 51.53\% \\
            & Classifier-based & - & 42.37\% & \textbf{77.04\%} & 68.72\% & \underline{72.96\%} & NaN & NaN & 52.22\% & 61.94\% & \textbf{62.54\%} \\
            \bottomrule
        \end{tabular}
        \begin{tablenotes}
            \footnotesize
            \item \textit{Note}: - indicates that this hyper-parameter is invalid for the current test; n-shot indicates that n samples will be provided for each prediction; k-shot provides a selection of k samples under each label to generate grimoire; r represents the sampling ratio of hard samples. The best performance in each column will be bolded, and the second-best performance will be underlined. NaN indicates that the number of valid experimental data is too small to give a reliable accuracy rate.
        \end{tablenotes}
    \end{threeparttable}
    }
\end{table*}

\begin{table*}[!t]
    \centering
    \resizebox{\linewidth}{!}{
    \begin{threeparttable}
        \caption{Demo of ten types of grimoire for the hate\_sp18 dataset.}
        \label{tab:grimoire demo}
        \begin{tabular}{cl}
            \toprule
            Grimoire type & \multicolumn{1}{c}{Grimoire text} \\
            \midrule
            \makecell[c]{KCS-PG \\ (k=4)} & \begin{tabular}[c]{@{}l@{}}Below are some skills needed to solve the task; you need to carefully learn and consider the process and methods step by step:\\ \\ To classify sentences into "hate" or "no hate," follow these rules:\\ \\ 1. **Identify Targeted Groups or Individuals:**\\    - Look for mentions of specific groups based on race, ethnicity, religion, gender, etc.\\    - Example: "This disgusting negro..." targets an individual based on race.\\    - Method: Scan the sentence for keywords that indicate a group or individual is being singled out.\\ \\ 2. **Assess Negative Language or Sentiments:**\\    - Determine if the sentence contains derogatory terms, insults, or negative stereotypes.\\    - Example: "...makes us (white people) look bad." uses a derogatory term and casts a negative stereotype.\\    - Method: Evaluate the adjectives and verbs used for negative connotations.\\ \\ 3. **Check for Exclusionary or Threatening Language:**\\    - Look for language that suggests exclusion, superiority, or threats towards a group.\\    - Example: "...vile hordes invading our lands..." implies a threat and exclusion.\\    - Method: Identify phrases that suggest conflict, invasion, or dehumanization.\\ \\ 4. **Analyze Context for Subtle Hate:**\\    - Sometimes hate is not overt; check for subtle implications or dog whistles.\\    - Example: "No need to allow FACTS to get in the way of a good White guilt message!" implies a subtle attack on a group.\\    - Method: Consider the broader context and potential implications of the sentence.\\ \\ 5. **Determine the Overall Message:**\\    - Evaluate if the sentence's overall message is meant to harm, demean, or incite negativity towards a group.\\    - Example: "That's their real contribution to society, misery and dependence." suggests a harmful stereotype.\\    - Method: Summarize the sentence's intent and check if it aligns with promoting hate.\\ \\ If a sentence meets any of the criteria for negative language, targeted attacks, or exclusionary sentiments towards a group or  individual, classify it as "hate." \\ Otherwise, classify it as "no hate."\end{tabular} \\
            \midrule
            \makecell[c]{KCS-SG \\ (k=4)} & \begin{tabular}[c]{@{}l@{}}Below are some skills needed to solve the task; you need to carefully learn and consider the process and methods step by step:\\ \\ 1. Look for specific groups or individuals being targeted based on identity like race or gender.\\ 2. Check if there are insults, negative stereotypes, or threatening language.\\ 3. Consider the overall intent, including subtle hate or broader context.\\ \\ Classify as "hate" if any criteria of targeting, negativity, or exclusion are met. If not, classify as "no hate."\end{tabular} \\
            \midrule
            \makecell[c]{HCS-PG \\ (k=4)} & \begin{tabular}[c]{@{}l@{}}Below are some skills needed to solve the task; you need to carefully learn and consider the process and methods step by step:\\ \\ To classify a sentence into "hate" or "no hate," follow these rules:\\ \\ 1. **Identify Hate Speech Indicators:**\\    - Look for derogatory language or slurs targeting a specific group based on race, ethnicity, religion, gender, sexual orientation, etc.\\    - Example: The use of "shegroid" is a derogatory term targeting a racial group, indicating hate speech.\\ \\ 2. **Assess Context and Intent:**\\    - Determine if the sentence promotes hostility, discrimination, or violence against a group.\\    - Example: "And they will flood into all white countries as much as they can!" suggests an invasion narrative against a particular racial group, indicating hate speech.\\ \\ 3. **Evaluate Tone and Content:**\\    - Analyze the tone for aggression or contempt towards a group.\\    - Example: "a nice kick to the nuts will always work... a black does n't equate to the same standard" shows contempt and dehumanization, indicating hate speech.\\ \\ 4. **Check for Advocacy of Harm or Exclusion:**\\    - Identify if the sentence advocates for harm or exclusion of a group.\\    - Example: "Here is possibly the best county to live in when one considers the Mud Index." The "Mud Index" likely refers to a derogatory way of rating places based \\ on the presence of certain groups, indicating hate speech.\\ \\ 5. **Distinguish Between Hate Speech and Non-Hate Speech:**\\    - If the sentence lacks any indicators of hate speech as described above, classify it as "no hate."\\    - Example: "I was gonna suggest an old Marlin 783 but then I remembered that it 's a .22 Mag." This sentence does not contain any indicators of hate speech.\\ \\ **Method of Applying the Rule:**\\ - Read the sentence carefully.\\ - Look for the indicators of hate speech.\\ - Consider the context, tone, and content.\\ - Decide if the sentence advocates for harm or exclusion.\\ - Classify as "hate" if any indicators are present; otherwise, classify as "no hate."\\ \\ Keep the analysis concise and focused on the presence or absence of hate speech indicators.\end{tabular} \\
            \midrule
            \makecell[c]{HCS-SG \\ (k=4)} & \begin{tabular}[c]{@{}l@{}}Below are some skills needed to solve the task; you need to carefully learn and consider the process and methods step by step:\\ \\ 1. Look for slurs or derogatory language aimed at a group's identity (race, gender, etc.).\\ 2. Check if the sentence encourages hostility, discrimination, or violence against a group.\\ 3. If no hate speech signs are found, label it as "no hate."\end{tabular} \\
            \bottomrule
        \end{tabular}
        \begin{tablenotes}
            \footnotesize
            \item \textit{Note}: Continue to the next page.
        \end{tablenotes}
    \end{threeparttable}
    }
\end{table*}

\begin{table*}[!t]
    \centering
    \resizebox{\linewidth}{!}{
    \begin{threeparttable}
        \caption*{-- continued from previous page}
        \label{tab:grimoire demo 2}
        \begin{tabular}{cl}
            \toprule
            Grimoire type & \multicolumn{1}{c}{Grimoire text} \\
            \midrule
            \makecell[c]{HSS-PG \\ (k=4,r=1.0,\\GPT3.5-Turbo)} & \begin{tabular}[c]{@{}l@{}}Below are some skills needed to solve the task; you need to carefully learn and consider the process and methods step by step:\\ \\ Rules for Classifying Sentences into Hate/No Hate:\\ \\ 1. Identify Targeted Groups or Stereotypes:\\    - Look for language that targets a specific group based on race, ethnicity, nationality, or any other social category.\\    - Example: "reds are just SO naive!" targets a group and implies naivety, indicating hate.\\    - Method: Scan for keywords that refer to groups and assess if the context is negative or derogatory.\\ \\ 2. Look for Generalizations and Derogatory Language:\\    - Generalizations apply a characteristic to all members of a group without exception.\\    - Derogatory language includes insults, slurs, or any language that demeans a group.\\    - Example: "you can take the ape out of the jungle..." uses a derogatory metaphor for a group, indicating hate.\\    - Method: Identify phrases that oversimplify or degrade a group and evaluate their intent.\\ \\ 3. Assess Context for Offensiveness:\\    - Determine if the sentence is likely to offend or harm the dignity of any group.\\    - Example: "Many Asians will see something once and memorize it..." suggests a stereotype, indicating hate.\\    - Method: Consider the potential impact of the statement on the targeted group.\\ \\ 4. Distinguish Between Criticism and Hate Speech:\\    - Criticism is directed at actions or policies and does not target or stereotype a group.\\    - Hate speech is directed at a group and is inherently derogatory or demeaning.\\    - Example: "Wow this policy of prosecuting people for offending liberals is disgusting." criticizes a policy, not a group, indicating no hate.\\    - Method: Focus on whether the criticism is aimed at behavior/policy or at a group's characteristics.\\ \\ 5. Evaluate Tone and Intent:\\    - Tone can indicate whether a statement is meant to be hateful or simply a blunt observation.\\    - Intent refers to whether the speaker aims to spread hate or is making a neutral statement.\\    - Example: "Sinners of the past did including your own people who sold you like cattle." may be a historical reference without hate intent, indicating no hate.\\    - Method: Analyze the tone and look for cues of malicious intent.\\ \\ By applying these rules, you can classify sentences into "hate" or "no hate" based on the presence of targeted derogatory language, generalizations, context, and the \\ distinction between criticism and hate speech.\end{tabular} \\
            \midrule
            \makecell[c]{HSS-SG \\ (k=4,r=1.0,\\GPT3.5-Turbo)} & \begin{tabular}[c]{@{}l@{}}Below are some skills needed to solve the task; you need to carefully learn and consider the process and methods step by step:\\ \\ 1. Check if the sentence targets a group negatively or uses stereotypes.\\ 2. Determine if the language is offensive or generalizes a group.\\ 3. Decide if it's criticizing behavior/policies or attacking a group's traits.\end{tabular} \\
            \midrule
            \makecell[c]{HSS-PG \\ (k=4,r=0.5,\\GPT3.5-Turbo)} & \begin{tabular}[c]{@{}l@{}}Below are some skills needed to solve the task; you need to carefully learn and consider the process and methods step by step:\\ \\ Rules for Classifying Sentences into Hate/No Hate:\\ \\ 1. Identify Discriminatory Language:\\    - Look for words or phrases that target a group based on race, ethnicity, gender, religion, or other social categories.\\    - Example: "mr whitey would be branded a big bad evil racist" targets a racial group.\\    - Method: Scan the sentence for keywords that indicate bias or stereotypes.\\ \\ 2. Assess Context for Negative Stereotypes:\\    - Determine if the sentence perpetuates negative stereotypes or expresses contempt.\\    - Example: "the little shops were all white runned and owned" implies a negative change related to racial demographics.\\    - Method: Consider the overall message and whether it implies harm or negativity towards a specific group.\\ \\ 3. Look for Generalizations or Assumptions:\\    - Check if the sentence makes broad generalizations about a group.\\    - Example: "reds are just SO naive" makes a sweeping assumption about a group.\\    - Method: Identify phrases that suggest all members of a group share a negative trait.\\ \\ 4. Evaluate Tone and Intent:\\    - Consider if the sentence has an aggressive or hostile tone.\\    - Example: "South Africa has returned to the Stone Age in 20 years" uses a derogatory comparison.\\    - Method: Read the sentence for emotional cues that suggest hate or aggression.\\ \\ 5. Distinguish Between Hate and Criticism:\\    - Differentiate between sentences that express hate and those that offer criticism without targeting a group.\\    - Example: "Wow this policy of prosecuting people for offending liberals is disgusting" criticizes a policy, not a group.\\    - Method: Focus on whether the criticism is directed at actions/ideas or at people/groups.\\ \\ 6. Consider Historical and Social Context:\\    - Be aware of historical and social connotations that may influence the interpretation of the sentence.\\    - Example: "Sinners of the past did including your own people who sold you like cattle" references history without expressing hate.\\    - Method: Use knowledge of social and historical contexts to inform your judgment.\\ \\ Apply these rules by carefully reading each sentence, identifying any elements that match the criteria for hate speech, and using your judgment to classify the \\ sentence accordingly. Remember to keep the analysis concise and focused on the key elements that indicate hate speech.\end{tabular} \\
            \midrule
            \makecell[c]{HSS-SG \\ (k=4,r=0.5,\\GPT3.5-Turbo)} & \begin{tabular}[c]{@{}l@{}}Below are some skills needed to solve the task; you need to carefully learn and consider the process and methods step by step:\\ \\ 1. Check for biased language targeting a group's race, gender, religion, etc.\\ 2. Look for negative stereotypes or generalizations about a group.\\ 3. Differentiate between hateful statements and criticism of ideas or actions.\end{tabular} \\
            \bottomrule
        \end{tabular}
        \begin{tablenotes}
            \footnotesize
            \item \textit{Note}: Continue to the next page.
        \end{tablenotes}
    \end{threeparttable}
    }
\end{table*}

\begin{table*}[!t]
    \centering
    \resizebox{\linewidth}{!}{
    \begin{threeparttable}
        \caption*{-- continued from previous page}
        \label{tab:grimoire demo 3}
        \begin{tabular}{cl}
            \toprule
            Grimoire type & \multicolumn{1}{c}{Grimoire text} \\
            \midrule
            \makecell[c]{RSS-PG \\ (k=4)} & \begin{tabular}[c]{@{}l@{}}Below are some skills needed to solve the task; you need to carefully learn and consider the process and methods step by step:\\ \\ Rules for Classifying Sentences into Hate/No Hate:\\ \\ 1. **Identify Targeted Groups or Individuals**: Look for references to specific groups or individuals based on race, ethnicity, religion, gender, or other identity \\ markers. Examples include "filthy jews" or "magic negro."\\ \\    *Method*: Scan the sentence for keywords that indicate a group or individual is being targeted. If such words are present, consider the context to determine if \\ they are used pejoratively.\\ \\ 2. **Assess the Tone and Intent**: Determine if the sentence has a derogatory, hostile, or demeaning tone towards the group or individual identified.\\ \\    *Method*: Read the sentence to gauge the emotional tone. Look for adjectives, adverbs, and verbs that convey negativity or hostility.\\ \\ 3. **Check for Generalizations and Stereotypes**: Sentences that make sweeping negative generalizations or stereotypes about a group are likely to be classified \\ as hate.\\ \\    *Method*: Look for statements that attribute negative characteristics or behaviors to an entire group rather than individuals.\\ \\ 4. **Evaluate Contextual Clues**: Sometimes, the context can shift a sentence from hate to no hate. For example, historical references or discussions without \\ negative intent may not be classified as hate.\\ \\    *Method*: Consider the broader context of the sentence. If the sentence is part of a historical discussion or lacks negative intent, it may not be classified as hate.\\ \\ 5. **Consider the Presence of Threats or Calls to Action**: Sentences that contain threats or calls to action against a group or individual are indicative of hate.\\ \\    *Method*: Look for verbs that suggest action or harm directed at a group or individual.\\ \\ By applying these rules, you can classify sentences into "hate" or "no hate." Remember that the context is crucial, and some sentences may require a nuanced \\ understanding of language and intent.\end{tabular} \\
            \midrule
            \makecell[c]{RSS-SG \\ (k=4)} & \begin{tabular}[c]{@{}l@{}}Below are some skills needed to solve the task; you need to carefully learn and consider the process and methods step by step:\\ \\ 1. Look for negative words about specific groups or people based on identity like race or gender.\\ 2. Check if the sentence sounds hostile or demeaning.\\ 3. Watch for broad negative statements about a whole group.\end{tabular} \\
            \midrule
            Zero-Shot-PG & \begin{tabular}[c]{@{}l@{}}Below are some skills needed to solve the task; you need to carefully learn and consider the process and methods step by step:\\ \\ To classify sentences into "hate" or "no hate," follow these summarized rules:\\ \\ 1. **Identify Hate Speech Indicators:**\\    - Look for words or phrases that target a group based on race, ethnicity, gender, religion, sexual orientation, disability, etc.\\    - Example: "All {[}group{]} are criminals" is hate speech.\\    - Method: Scan the sentence for derogatory terms or generalizations about a group.\\ \\ 2. **Assess Context and Tone:**\\    - Determine if the language is aggressive, disrespectful, or intended to harm.\\    - Example: "I hate {[}group{]}" indicates hate speech, while "I hate broccoli" does not.\\    - Method: Consider the sentence's context and the potential impact on the targeted group.\\ \\ 3. **Look for Incitement to Violence:**\\    - Check if the sentence encourages harm or violence against individuals or groups.\\    - Example: "Someone should teach {[}group{]} a lesson" implies hate speech.\\    - Method: Identify calls to action that suggest physical harm or intimidation.\\ \\ 4. **Evaluate for Exclusion or Discrimination:**\\    - Determine if the sentence promotes exclusion or discrimination against a protected group.\\    - Example: "No {[}group{]} allowed" is hate speech.\\    - Method: Look for statements that advocate for unequal treatment or segregation.\\ \\ 5. **Consider Intent and Sarcasm:**\\    - Be aware that sarcasm or satire may complicate classification.\\    - Example: "Yeah, right, like all {[}group{]} are geniuses" could be sarcastic.\\    - Method: Use contextual clues to discern the speaker's true intent.\\ \\ 6. **Use Reliable Classification Tools:**\\    - If available, employ automated tools or guidelines provided by reputable organizations for consistency.\\    - Example: AI-based text analysis tools can help classify sentences.\\    - Method: Input the sentence into the tool and review the classification, using human judgment as necessary.\\ \\ Remember, the classification should be based on the content of the sentence and not personal opinions or biases. If a sentence does not clearly fall into the \\ "hate" category, it should be classified as "no hate."\end{tabular} \\
            \midrule
            Zero-Shot-SG & \begin{tabular}[c]{@{}l@{}}Below are some skills needed to solve the task; you need to carefully learn and consider the process and methods step by step:\\ \\ 1. Check for hate speech: Look for words that insult or generalize about race, gender, religion, etc. Consider if the sentence is aggressive or aims to harm.\\ 2. Check for violence or discrimination: See if the sentence encourages violence or unfair treatment against certain groups.\\ 3. Use context and tools: Pay attention to sarcasm and context. Use reliable tools to help classify if needed. If unsure, label as "no hate."\end{tabular} \\
            \bottomrule
        \end{tabular}
        \begin{tablenotes}
            \footnotesize
            \item \textit{Note}: Continue to the next page.
        \end{tablenotes}
    \end{threeparttable}
    }
\end{table*}

\end{document}